\documentclass[sn-nature]{sn-jnl}

\usepackage{graphicx}%
\usepackage{multirow}%
\usepackage{amsmath,amssymb,amsfonts}%
\usepackage{amsthm}%
\usepackage{mathrsfs}%
\usepackage[title]{appendix}%
\usepackage{xcolor}%
\usepackage{textcomp}%
\usepackage{manyfoot}%
\usepackage{booktabs}%
\usepackage{algorithm}%
\usepackage{algorithmicx}%
\usepackage{algpseudocode}%
\usepackage{listings}%
\usepackage{fontawesome5}
\usepackage{makecell}
\usepackage{caption}
\usepackage{hyperref}
\usepackage{marginnote}
\usepackage{bbding}
\usepackage{pifont}
\usepackage{subcaption}
\usepackage{float}
\usepackage{lineno}
\usepackage{array}
\usepackage{colortbl}
\usepackage[export]{adjustbox}
\hypersetup{
    colorlinks=true,
    urlcolor=black,
}

\usepackage{changes}

\setaddedmarkup{\textcolor{blue}{#1}}  
\setdeletedmarkup{}  

\newcolumntype{C}{>{\color{blue}}c}
\newcolumntype{L}{>{\color{blue}}l}

\begin{document}
\title[Article Title]{A multi-modal vision-language model for generalizable annotation-free pathology localization}


\author[1,2,3]{\fnm{Hao} \sur{Yang}}
\equalcont{These authors contributed equally to this work.}
\author[4]{\fnm{Hong-Yu} \sur{Zhou}}
\equalcont{These authors contributed equally to this work.}
\author[1,2,3]{\fnm{Jiarun} \sur{Liu}}
\author[1,2,3]{\fnm{Weijian} \sur{Huang}}
\author[1]{\fnm{Cheng} \sur{Li}}
\author[5,6]{\fnm{Zhihuan} \sur{Li}}
\author[5]{\fnm{Yuanxu} \sur{Gao}}
\author[7]{\fnm{Qiegen} \sur{Liu}}
\author[2,8]{\fnm{Yong} \sur{Liang}}
\author[9,10]{\fnm{Qi} \sur{Yang}}
\author[11]{\fnm{Song} \sur{Wu}}
\author[12]{\fnm{Tao} \sur{Tan}}
\author[1]{\fnm{Hairong} \sur{Zheng}}
\author*[5,6]{\fnm{Kang} \sur{Zhang}}
\author*[1]{\fnm{Shanshan} \sur{Wang}}
\email{ss.wang@siat.ac.cn}
\email{kang.zhang@gmail.com}

\affil[1]{\orgdiv{Paul C. Lauterbur Research Center for Biomedical Imaging}, \orgname{Shenzhen Institutes of Advanced Technology, Chinese Academy of Sciences}, \city{Shenzhen}, \country{China}}
\affil[2]{\orgname{Pengcheng Laboratory}, \city{Shenzhen}, \country{China}}
\affil[3]{\orgname{University of Chinese Academy of Sciences}, \city{Beijing}, \country{China}}
\affil[4]{\orgdiv{School of Biomedical Engineering}, \orgname{Tsinghua University}, \city{Beijing}, \country{China}}
\affil[5]{\orgname{Institute for AI in Medicine and Faculty of Medicine}, \orgname{Macau University of Science and Technology}, \city{Macau}, \country{China}}
\affil[6]{\orgdiv{State Key Laboratory of Eye Health, Eye Hospital and Institute for Advanced Study on Eye Health and Diseases, Institute for Clinical Data Science}, \orgname{Wenzhou Medical University}, \city{Wenzhou}, \country{China}}
\affil[7]{\orgdiv{Department of Electronic Information Engineering}, \orgname{Nanchang University}, \city{Nanchang}, \country{China}}
\affil[8]{\orgdiv{Chinese Medicine Guangdong Laboratory}, \city{Hengqin}, \country{China}}
\affil[9]{\orgname{Beijing Chaoyang Hospital, Capital Medical University}, \city{Beijing}, \country{China}}
\affil[10]{\orgdiv{Key Lab of Medical Engineering for Cardiovascular Disease}, \orgname{Ministry of Education}, \city{Beijing}, \country{China}}
\affil[11]{\orgdiv{Department of Urology, South China Hospital}, \orgname{Medical School, Shenzhen University}, \city{Shenzhen}, \country{China}}
\affil[12]{\orgdiv{Faculty of Applied Sciences}, \orgname{Macao Polytechnic University}, \city{Macau}, \country{China}}

\abstract{
Existing deep learning models for defining pathology from clinical imaging data rely on expert annotations and lack generalization capabilities in open clinical environments. Here, we present a generalizable vision-language model for Annotation-Free pathology Localization (AFLoc). The core strength of AFLoc is extensive multi-level semantic structure-based contrastive learning, which comprehensively aligns multi-granularity medical concepts with abundant image features to adapt to the diverse expressions of pathologies without the reliance on expert image annotations. We conduct primary experiments on a dataset of 220K pairs of image-report chest X-ray images and perform validation across eight external datasets encompassing 34 types of chest pathologies. The results demonstrate that AFLoc outperforms state-of-the-art methods in both annotation-free localization and classification tasks. Additionally, we assess the generalizability of AFLoc on other modalities, including histopathology and retinal fundus images. We show that AFLoc exhibits robust generalization capabilities, even surpassing human benchmarks in localizing five different types of pathological images. These results highlight the potential of AFLoc in reducing annotation requirements and its applicability in complex clinical environments.}


\maketitle

\begin{figure}[bht]
  \centering
  \centerline{\includegraphics[width=17cm]{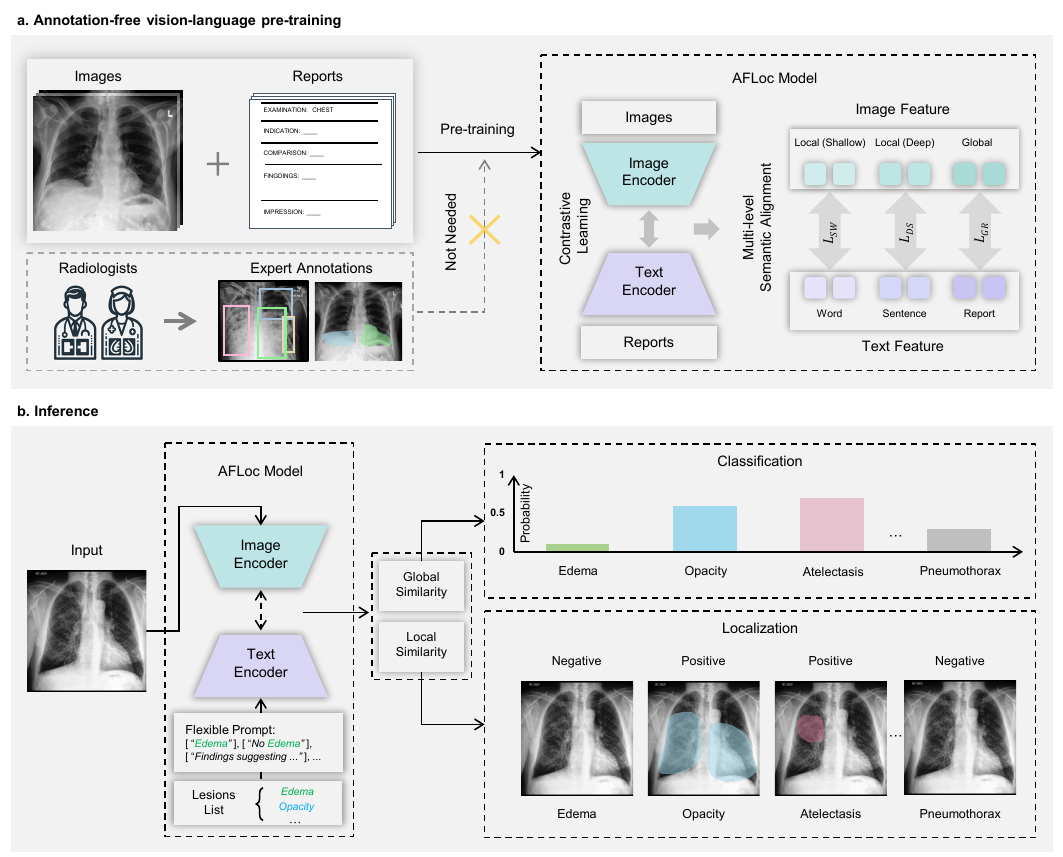}}
  \caption{Overview of the AFLoc’s annotation-free pipeline for pathology localization. \textbf{a.} Annotation-free vision-language pre-training: AFLoc leverages contrastive-based vision-language pre-training with existing images and text reports to eliminate the need for additional annotation efforts. A multi-level semantic alignment scheme is proposed to facilitate the comprehensive alignment of medical concepts across text reports with image features. \textbf{b.} Inference: AFLoc can classify and localize all potential pathologies within the lesion list. The model encodes the input image and the automatically generated text prompts into feature embeddings. Then the local and global level feature similarity are computed for localization and classification respectively. The model only output localization results with positive classification predictions.} 
  \medskip
  \label{fig:method}
\end{figure}
Accurate diagnosis and precise pathology localization in medical images facilitate customized treatment approaches that improve patient outcomes and mitigate the possibility of diagnostic errors. By pinpointing the exact location and extent of abnormalities, clinicians can make informed decisions that lead to more targeted therapies and improved prognoses for patients \cite{coudray2018classification,wang2018development,campanella2019clinical,courtiol2019deep}. 

Over the past decade, supervised deep learning methods have accelerated advancements in disease localization \cite{lee2022localization,cao2023large,song2020clinically,leon2023hyperspectral}. However, the efficacy of these methods heavily relies on extensively annotated training datasets, which require domain experts to invest considerable time \cite{rajpurkar2022ai,wang_annotation-efficient_2021}. Specifically, clinical localization tasks often require experienced clinicians to meticulously annotate numerous precise bounding boxes or perform pixel-wise delineations of localized pathology areas. This annotation process is costly, particularly in resource-constrained clinical settings, and algorithms frequently struggle to generalize to diverse datasets.

Several methods have been proposed to reduce the reliance on large annotated datasets \cite{wang_annotation-efficient_2021,liu2024swin,chen2020simple,zhou2023foundation}. Initially, these methodologies acquire general visual representations through self-supervised learning from image datasets, followed by fine-tuning on smaller annotated datasets. This approach enables models to achieve high performance on specific tasks while decreasing the need and cost of data labeling \cite{zhou2023foundation}. Moreover, saliency-based methods \cite{saporta2022benchmarking,gradcam,gradcam++,eigencam} have been developed to reduce annotation costs in pathology localization tasks by allowing coarse localization of target categories in models trained with image-level annotations. However, these methods still require annotations for specific downstream tasks. This requirement is particularly challenging in flexible and dynamic clinical environments, especially for emerging diseases (e.g., COVID-19), where deployed models may fail to perform effectively \cite{tiu2022expert,wu2023medklip, morens2013emerging}.

In recent years, unsupervised deep learning methods have gained increasing attention due to their independence from annotated datasets, particularly in the field of anomaly detection \cite{baur2021autoencoders,rd,rd++,ReContrast}. These methods typically train models using only healthy samples, learning the distribution of normal anatomical structures, which enables the identification of abnormal pathology samples during the testing phase \cite{baur2021autoencoders,rd}. They are particularly effective for data with simple structures and low inter-sample variance, allowing them to learn normative distributions and achieve excellent anomaly detection performance \cite{rd++,ReContrast,upd}. However, challenges such as the high heterogeneity of pathology images, similarities between different pathologies, and large variations in contrast for the same lesions reduce the usability of these methods in complex scenarios, thereby hindering their practical application in real medical environments \cite{upd}.

A promising approach is the development of medical visual-language pre-training methods \cite{huang_nc_2024,huang_mia_2024,boecking2022making,convirt,wu2023medklip,huang2021gloria,zhou2023advancing}. These methods establish effective correlations between medical reports and medical images, allowing them to flexibly localize disease types not encountered during pre-training without requiring additional customized annotations \cite{wu2023medklip}. However, achieving precise pathology localization solely through the combination of medical images and clinical reports remains challenging. A primary obstacle is the lack of explicit pathology localization markers in clinical reports, which often provide only coarse information such as `upper' or `left' to indicate disease location. Moreover, clinical descriptions by clinicians are subjective and variable, further complicating the task of accurately extracting and localizing diseases in medical images. To address this challenge, several methods have been proposed to integrate finer-grained information. For instance, GLoRIA \cite{huang2021gloria} extracts the correlation of the image's regions and paired words in reports to learn global and local representations of images. MedKLIP \cite{wu2023medklip} uses well-defined medical vocabulary knowledge bases to provide supervision at the entity level through triplet training paradigms. However, these fine-grained methods typically focus on individual levels of medical concepts and may overlook the variable meanings of concepts in different contexts. Therefore, these approaches may struggle to adapt to the diverse expressions of disease descriptors in clinical practice, often requiring customized textual cues to enhance localization performance.

In this study, we propose AFLoc, a visual-language model based on contrastive learning aimed at alleviating the need for costly pathology localization annotations. AFLoc can autonomously perform pathology localization and clinical diagnosis with medical images. Unlike traditional global semantic alignment strategies \cite{convirt,boecking2022making}, AFLoc introduces a contrastive learning framework with a multi-level semantic alignment component, facilitating the comprehensive alignment of medical concepts from reports with image features. Specifically, the image encoder generates three levels of features: shallow local features, deep local features, and global features, which are aligned with word-level, sentence-level, and report-level features extracted by the text encoder. We extensively validated AFLoc across three types of medical image datasets, including chest X-ray (8 external datasets), histopathology (3 external datasets), and retinal fundus images. Our results show that AFLoc outperforms state-of-the-art methods in localization and clinical diagnostic tasks across different modalities. We hope that this study can help address the challenges posed by annotation scarcity and modality diversity in clinical environments, while providing insights for the design of future clinical open-environment methods.

\begin{figure}[bht]
    \centering
    \centerline{\includegraphics[width=15cm]{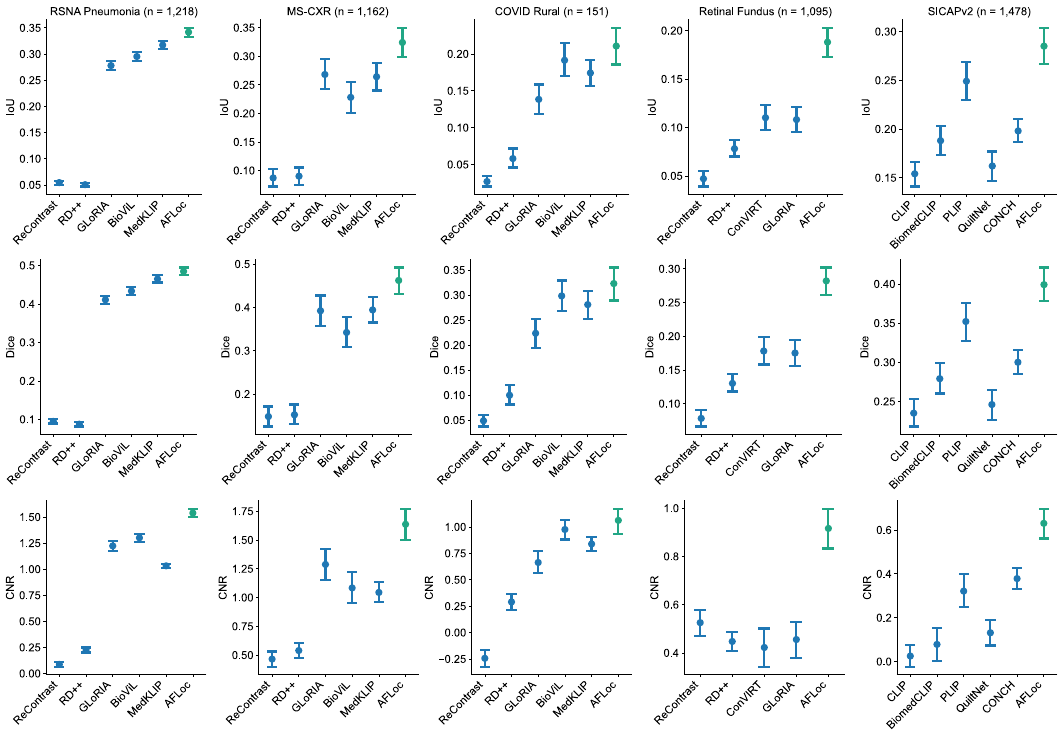}}
    \caption{
    Comparisons of AFLoc with state-of-the-art methods across five evaluation datasets for pathology localization in chest X-ray, retinal fundus, and histopathology images. For each method–dataset pair, IoU, Dice similarity coefficient, and CNR are reported. The central dots represent the mean, and the vertical error bars indicate the 95\% CI. The variable n denotes the number of evaluation images in each dataset. Detailed results are provided in Supplementary Table\ref{extab:loc_cxr3}-\ref{extab:loc_path}.}
    \label{fig:loc}
\end{figure}

\begin{figure}[bht]
  \centering
    \centerline{\includegraphics[width=15cm]{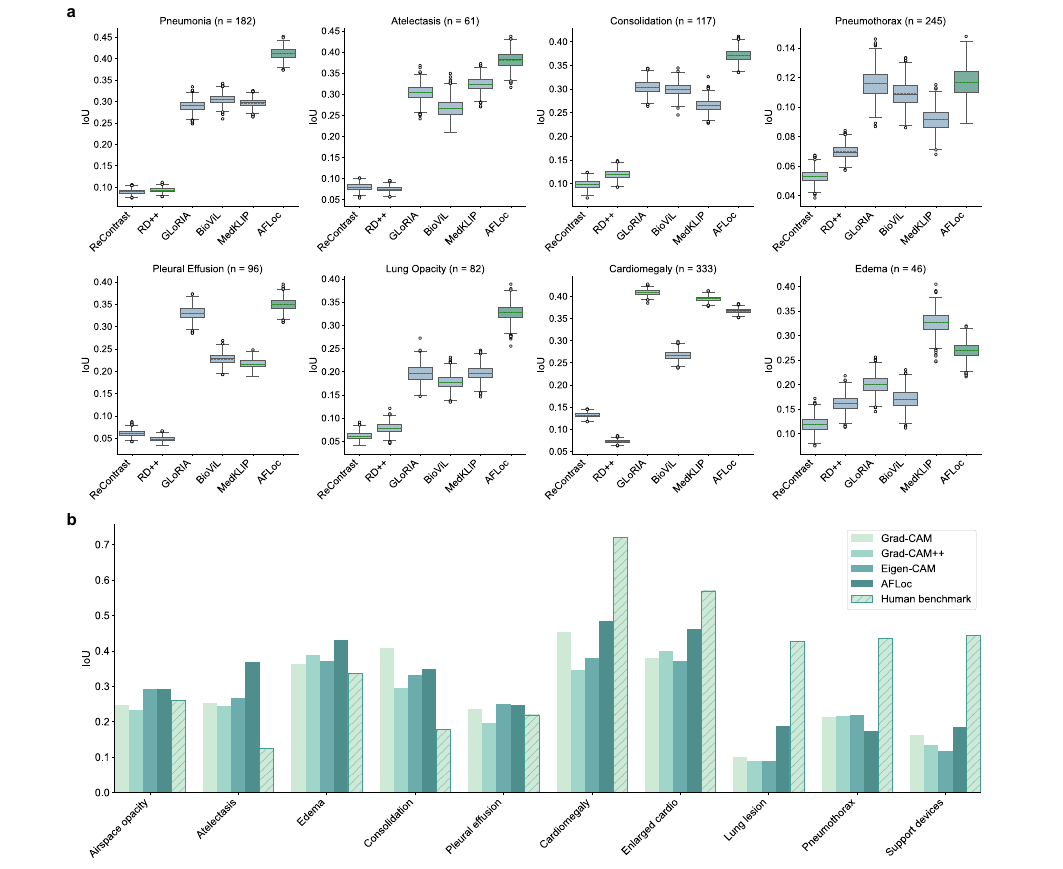}}
  \caption{
  Comparisons of AFLoc in the task of pathology localization across different chest pathologies. \textbf{a}. Results on the MS-CXR dataset, compared with existing state-of-the-art unsupervised anomaly detection models and vision-language models. The variable n denotes the number of evaluation images in each dataset. In each boxplot, the solid center line represents the median, the dashed line represents the mean, the box boundaries correspond to the first and third quartiles, the whiskers extend to the most extreme data points that are not considered outliers, and the outliers are represented by dots. \textbf{b}. Results on the CheXlocalize dataset, compared with various saliency methods and the human benchmark.}
  \medskip
  \label{fig:loc_cxr}
\end{figure}

\begin{figure}[bht]
  \centering
  \centerline{\includegraphics[width=15cm]{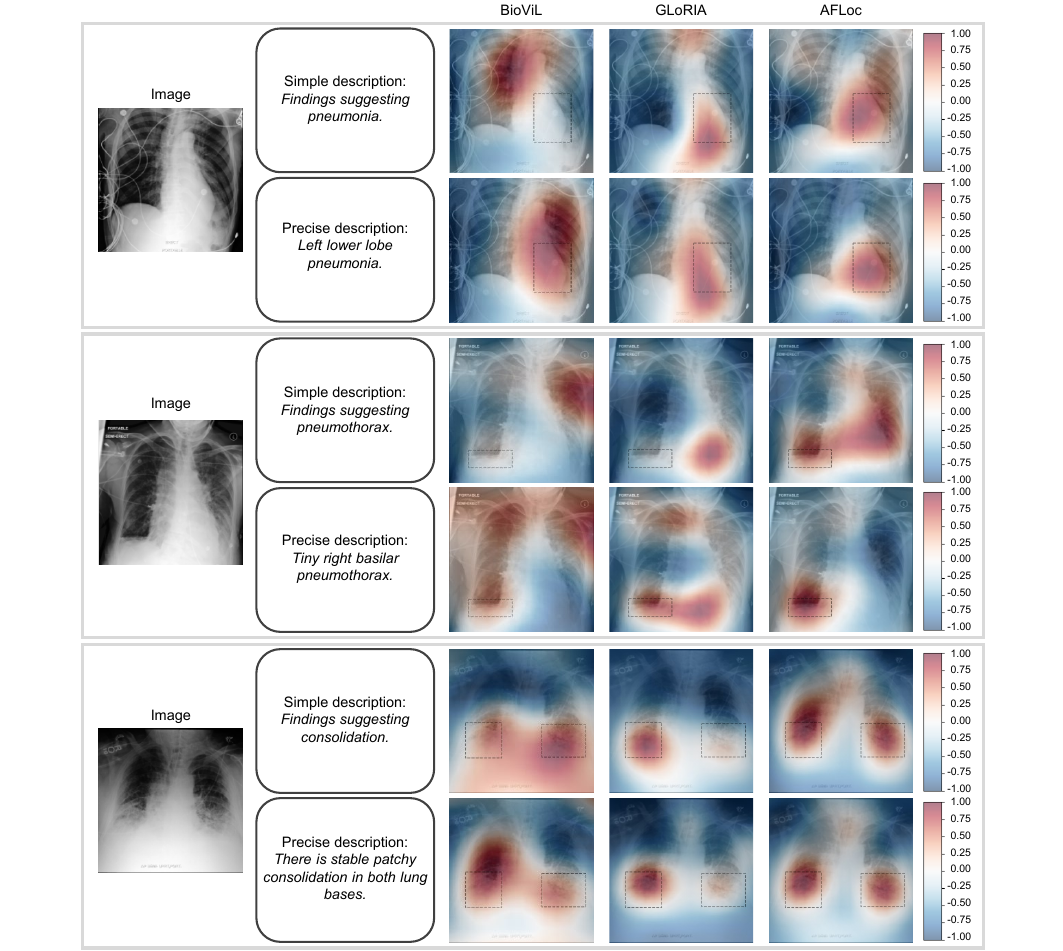}}
  \caption{Visualization of pathology localization results across different imaging modalities and diseases. Black dashed boxes indicate the pathology areas marked by radiologists, and the intensity of red color in the heatmaps signifies the focus level of the model's prediction, with deeper red indicating higher focus.}
  \medskip
  \label{fig:loc_vis}
\end{figure}

\section*{Results}\label{sec2}
We primarily evaluated AFLoc on two tasks: pathology localization and clinical diagnosis. Three modalities were considered, including chest X-ray, histopathology, and retinal fundus images. In addition, we conducted extensive ablation studies to evaluate the effects of different prompts, text granularities, and text encoders on AFLoc's performance. These experiments are discussed in detail in subsequent sections.


\subsubsection*{Annotation-Free Localization Across Three Medical Modalities}\label{sec3}
AFLoc is capable of performing zero-shot localization tasks without requiring any annotations. In this section, we present AFLoc's annotation-free localization performance across different clinical modalities. Each modality has been pre-trained using its corresponding image-report datasets, ensuring modality-specific knowledge is effectively captured and used. Here, we consider that 'pathology localization' encompasses both heatmap predictions and binary mask results, as some clinical applications may benefit more from the visualization provided by heatmaps rather than discrete segmentation. We quantitatively evaluate each predicted heatmap using Contrast-to-Noise Ratio (CNR) \cite{boecking2022making} and assess the thresholded segmentation masks using Intersection over Union (IoU) \cite{rezatofighi2019generalized} and Dice similarity coefficient metrics. In the following, we will present the results of the three modalities separately.

\textbf{Localization Performance on Chest X-Ray}
In clinical practice for chest diseases, doctors often need to perform detailed assessments of chest X-rays for early screening and surgical planning. For instance, airspace opacity can indicate lung infections, inflammation, or tumors \cite{reed2011multifocal}. Atelectasis leads to restricted gas exchange, causing hypoxemia and severely affecting respiratory function \cite{woodring1996types}. However, since these diseases often lack clear boundaries and have contrast levels similar to surrounding tissues, doctors typically need to invest significant effort in distinguishing them. Therefore, automated diagnostic systems play a crucial role in aiding diagnosis.

We evaluated the localization performance of AFLoc on chest X-rays using four external datasets: RSNA Pneumonia \cite{shih2019augmenting}, MS-CXR \cite{boecking2022making}, CheXlocalize \cite{saporta2022benchmarking}, and COVID Rural \cite{desai2020chest,tang2020deep}. These datasets cover 13 common thoracic pathologies: pneumonia, airspace opacity, atelectasis, cardiomegaly, consolidation, edema, enlarged cardiomediastinum, lung lesion, pleural effusion, pneumothorax, lung opacity, support devices, and COVID-19.

We first elaborate AFLoc's localization performance on the RSNA Pneumonia dataset (Fig. \ref{fig:loc}, Supplementary Table\ref{extab:loc_cxr3}). AFLoc achieves better results than all comparative methods across all three evaluation metrics. 
Specifically, compared to the respective best-performing comparative methods, AFLoc boosts the IoU by 7.9\% from 0.317 (95\% CI: 0.309, 0.325) to 0.342 (95\% CI: 0.332, 0.350), enhances the Dice coefficient by 4.1\% from 0.465 (95\% CI: 0.455, 0.474) to 0.484 (95\% CI: 0.474, 0.495), and elevates CNR by 18.3\% from 1.300 (95\% CI: 1.262, 1.339) to 1.538 (95\% CI: 1.496, 1.580).
These results highlight AFLoc's superior ability to localize pathological regions.

Similarly, the results on the MS-CXR dataset (Fig. \ref{fig:loc}, Supplementary Table\ref {extab:loc_cxr3} and \ref{extab:loc_mscxr}) show that our AFLoc outperforms the existing vision-language pre-training approaches (GLoRIA, BioViL, and MedKLIP) by a clear margin. 
For example, AFLoc improves the IoU (0.324, 95\% CI: 0.298, 0.350) by more than 6\% compared to GLoRIA (0.268, 95\% CI: 0.242, 0.295), BioViL (0.228, 95\% CI: 0.201, 0.255), and MedKLIP (0.264, 95\% CI: 0.240, 0.288). AFLoc shows consistent performance enhancement across the majority of pathologies (six out of eight) (Fig. \ref{fig:loc_cxr}a), confirming its superior pathology localization capability.
In addition, on the CheXlocalize dataset, we compared AFLoc with saliency-based methods (Grad-CAM, Grad-CAM++, and Eigen-CAM) and the human benchmark using results reported in the literature \cite{saporta2022benchmarking} (Fig. \ref{fig:loc_cxr}b). AFLoc outperforms these methods in most cases, achieving a 4\% higher average IoU than Grad-CAM (0.318 vs. 0.282). AFLoc also surpasses the human benchmark in localizing five pathologies, including airspace opacity and pleural effusion, demonstrating its potential in medical image analysis.

Finally, we demonstrate AFLoc’s ability to generalize to unseen diseases by using the COVID Rural dataset (X-rays of COVID-19 pneumonia). AFLoc achieves an IoU of 0.211 (95\% CI: 0.185, 0.236), surpassing the performance of MedKLIP (0.174, 95\% CI: 0.156, 0.192). AFLoc also achieves higher Dice coefficient (0.323, 95\% CI: 0.289, 0.355) and CNR (1.062, 95\% CI: 0.929, 1.173) when compared to all comparative models. These results underscore AFLoc's potential for real-world applications of AFLoc in clinical scenarios with emerging conditions.

\textbf{Localization Performance on Retinal Fundus Images}
To further demonstrate the generalizability of our AFLoc, we applied it to retinal fundus images. We curated a retinal fundus image dataset containing three prevalent retinal pathologies: choroidal neovascularization (CNV), drusen, and intraretinal hemorrhages. Detecting choroidal neovascularization is essential for the early diagnosis of wet age-related macular degeneration (AMD) \cite{lim2012age}. Drusen serves as an indicator of dry AMD \cite{lim2012age}. Intraretinal hemorrhages can reflect severe conditions such as diabetic retinopathy or hypertensive retinopathy \cite{uhler2008optic}. Early detection and identification of these pathologies can significantly slow disease progression and facilitate proactive interventions.

The localization results of different methods on these three retinal pathologies are depicted in Fig. \ref{fig:loc} (column 4) and Supplementary Table\ref{extab:loc_fundus}. Notably, AFLoc attains the highest IoU, Dice, and CNR scores across the different pathologies. For choroidal neovascularization, AFLoc achieves an IoU of 0.321 (95\% CI: 0.303, 0.337), which is much higher than GLoRIA (0.143, 95\% CI: 0.130, 0.156) and other comparative methods. Regarding drusen, AFLoc achieves an IoU of 0.147 (95\% CI: 0.125, 0.170) and a Dice score of 0.225 (95\% CI: 0.192, 0.257), showcasing its proficiency in localizing subtle features. 

While AFLoc achieves superior scores compared to all comparative methods, its capability to localize intraretinal hemorrhages is relatively weaker than for the other two pathologies, with an IoU of 0.097 (95\% CI: 0.091, 0.102). This suggests the necessity for further enhancements in localizing diffuse and less distinct lesions. In Fig. \ref{fig:loc_vis}, the visualization of retinal localization is presented, with deeper red areas denoting model-predicted disease locations, while the black box signifies clinician annotations. AFLoc accurately localizes concentrated pathologies like choroidal neovascularization and drusen.

In summary, the localization results in this section underscore that AFLoc proves effective not only in chest X-ray images, as discussed in the previous section, but also in retinal fundus images. This showcases its potential as a versatile localization tool capable of generalizing across various imaging modalities.

\begin{figure}[bht]
  \centering
  \centerline{\includegraphics[width=13cm]{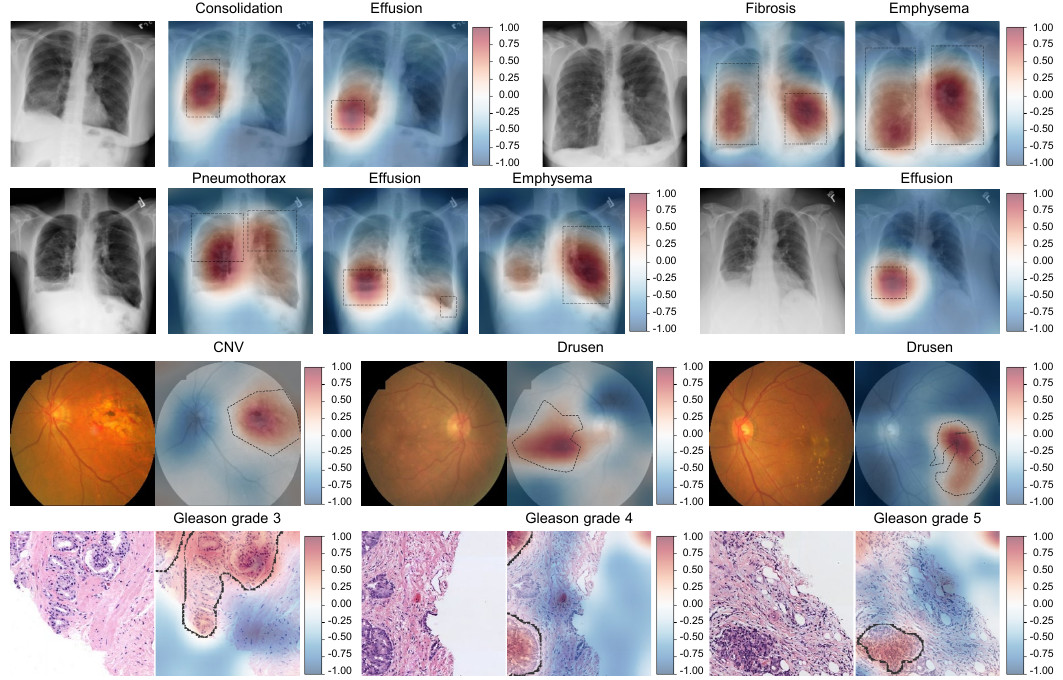}}
  \caption{Visualization of pathology localization by different models on MS-CXR. Results with simple and precise descriptions are shown to demonstrate the models' performance under different description granularities. Black dashed boxes indicate the pathology areas marked by radiologists, while deeper red indicates a higher focus level in the models' predictions.}
  \medskip
  \label{fig:two_desc}
\end{figure}

\textbf{Localization Performance on Histopathology}
In histopathology, accurate localization of abnormal tissues is key for cancer diagnosis and grading. For example, in prostate cancer, the Gleason grading system is used to assess tissue samples by examining glandular architecture, nuclear features, and lumen formation. Pathologists assign a Gleason score based on the tissue’s differentiation, reflecting the cancer's aggressiveness \cite{sicap}. Early detection and identification of these pathologies can significantly slow disease progression and facilitate preventive interventions.

Here, we used the Quilt-1M dataset \cite{quilt} to train the AFLoc model for histopathology and tested its performance on the SICAPv2 test set (n=2,122) under the annotation-free localization setting. We compared AFLoc with several state-of-the-art methods, including CLIP \cite{clip}, BiomedCLIP \cite{biomedclip}, and three foundation models specialized for histopathology: PLIP \cite{plip}, QuiltNet \cite{quilt}, and CONCH \cite{conch}. Specifically, we applied GradCAM \cite{gradcam} for the comparative methods, as it has been shown to enhance their performance.

As shown in Fig. \ref{fig:loc} and Supplementary Table\ref{extab:loc_path}, AFLoc achieves the best localization performance across all three evaluation metrics. 
Specifically, compared to the best-performing comparative method, PLIP, AFLoc demonstrates notable improvements in IoU (0.285, 95\% CI: 0.267, 0.304 vs. 0.249, 95\% CI: 0.230, 0.269), Dice score (0.399, 95\% CI: 0.378, 0.421 vs. 0.352, 95\% CI: 0.327, 0.376), and CNR (0.630, 95\% CI: 0.561, 0.697 vs. 0.321, 95\% CI: 0.249, 0.398).
These results suggest that AFLoc is capable of more accurately localizing abnormal tissues compared to both general-purpose and histopathology-specific models.

While the results are encouraging, it is worth noting that histopathological localization remains a highly challenging task due to the complexity and variability of tissue morphology. Nevertheless, AFLoc’s superior performance across all metrics signifies its potential as a robust and generalizable tool for annotation-free pathology localization in histopathology.

\subsubsection*{Ablation Study of Two Types of Prompts}
In the previous section, we validated AFLoc's ability for annotation-free localization across three imaging modalities using tailored prompts. In clinical settings, there are two prompt generation methodologies: 1) automated prompts based on general-purpose schemes for different imaging modalities, and 2) manually crafted prompts for enhanced precision in localization, as provided by the MS-CXR dataset. This raises the question: to what extent can customized prompts improve AFLoc's performance?

To explore this, we compared the results obtained using simple and precise prompts on the MS-CXR dataset. For example, the simple prompt for pneumonia can be `findings suggesting pneumonia', while the precise description can be `severe bibasilar consolidation' or `air space opacity in a right infrahilar location', reflecting clinical terminology.

Results in Table \ref{tab:two_disc} show that when using precise prompts, all models, including our AFLoc and the two comparative methods (BioViL and GLoRIA), demonstrate enhanced localization performance. For our AFLoc, the use of precise prompts leads to a 12.1\% increase in IoU (from 0.289 to 0.324), a 10.5\% increase in Dice (from 0.418 to 0.462), and a 21.1\% increase in CNR (from 1.351 to 1.636). These improvements highlight the advantages of detailed, clinically relevant prompts to enhance localization accuracy.
Nevertheless, AFLoc outperforms the two comparative methods regardless of the prompt generation methods. 

Fig. \ref{fig:two_desc} illustrates that with simple prompts, models tend to have broader, less accurate focal areas, frequently failing to align with radiologist-annotated regions. In contrast, with precise prompts, AFLoc exhibits enhanced focus, closely matching the annotations made by radiologists. For example, in the case of left lower lobe pneumonia, AFLoc accurately highlights the affected region with precise prompts. Similarly, for the case involving tiny right basilar pneumothorax, much better localization performance is observed when precise prompts are used compared to simple prompts. Moreover, we show the model’s ability to generate distinct localization responses when prompted with different disease descriptions, as shown in Extended Data Fig. \ref{exfig:overlap}. These examples include multi-lesion cases with varying degrees of spatial overlap. Notably, the model accurately localizes the corresponding abnormality based on the prompts with simple descriptions for different diseases.

\begin{table}[ht]   
    \centering
    \caption{
    \deleted{Quantitative results of different models on the MS-CXR dataset evaluated under two different descriptive scenarios. Numbers within parentheses indicate 95\% CI.}
    Comparisons of localization performance on the MS-CXR dataset with different descriptive granularities. Numbers within parentheses indicate 95\% CI.}
    \label{t1}
    \arrayrulecolor{black}
    \begin{tabular}{lccc}
        \toprule[1pt] 
                    \textbf{Description} & \textbf{BioViL} & \textbf{GLoRIA} & \textbf{AFLoc} \\ \hline
        \textbf{IoU}                 &        &        &       \\ \hline
        Simple description  & \makecell[l]{0.187\\(0.162,0.213)}  & \makecell[l]{0.240\\(0.214,0.265)}   & \makecell[l]{\textbf{0.289}\\\textbf{(0.262,0.314)}} \\
        Precise description & \makecell[l]{0.228\\(0.201,0.255)}  & \makecell[l]{0.268\\(0.242,0.295)}   & \makecell[l]{\textbf{0.324}\\\textbf{(0.298,0.350)}} \\ 
        \cmidrule[1pt]{1-4}
        \textbf{Dice}                  &        &        &       \\ \hline
        Simple description  & \makecell[l]{0.287\\(0.252,0.321)}  & \makecell[l]{0.357\\(0.322,0.392)}  & \makecell[l]{\textbf{0.418}\\\textbf{(0.383,0.450)}}  \\
        Precise description & \makecell[l]{0.342\\(0.308,0.377)}  & \makecell[l]{0.392\\(0.357,0.427)}  & \makecell[l]{\textbf{0.462}\\\textbf{(0.431,0.492)}}  \\ 
        \cmidrule[1pt]{1-4}
        \textbf{CNR}                  &        &        &       \\ \hline
        Simple description  & \makecell[l]{0.825\\(0.694,0.948)}  & \makecell[l]{1.128\\(1.003,1.254)}  & \makecell[l]{\textbf{1.351}\\\textbf{(1.219,1.481)}}  \\
        Precise description & \makecell[l]{1.083\\(0.949,1.219)}  & \makecell[l]{1.287\\(1.149,1.421)}  & \makecell[l]{\textbf{1.636}\\\textbf{(1.501,1.772)}} \\ 
        \bottomrule[1pt]
    \end{tabular}
    \label{tab:two_disc}
        \begin{tablenotes}
        \footnotesize
        \item Bold values represent the highest performance score among the compared methods.
    \end{tablenotes}

\end{table}

\subsubsection*{Annotation-Free Diagnosis Across Three Medical Modalities}

In this section, we demonstrate AFLoc's annotation-free diagnostic performance through zero-shot classification, using CXR, retinal fundus images, and histopathology datasets. For comparison, we also present the results of two unsupervised anomaly detection methods, ReContrast and RD++. These methods, without fine-tuning using labeled datasets, are referred to as `zero-shot' methods for convenience.



\begin{figure}[bht]
  \centering
  \centerline{\includegraphics[width=17cm]{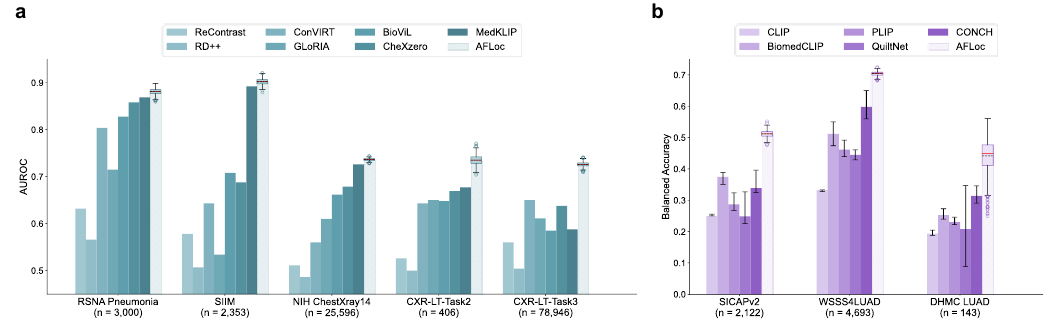}}
  \caption{Comparisons for the zero-shot classification task across chest X-ray and histopathology images. \textbf{a.} Results of AUROC on the RSNA Pneumonia, SIIM, NIH ChestXray14, and CXR-LT datasets. \textbf{b.} Results of balanced accuracy on the SICAPv2, WSSS4LUAD, and DHMC LUAD datasets.Comparison methods are displayed as bar plots with error bars denoting 95\% CI. Our method is shown as boxplots. The variable n denotes the number of evaluation images in each dataset. In each boxplot, the solid center line represents the median, the dashed line represents the mean, the box boundaries correspond to the first and third quartiles, the whiskers extend to the most extreme data points that are not considered outliers, and the outliers are represented by dots.}
  \medskip
  \label{fig:zeroshot_cls}
\end{figure}

\textbf{Diagnosis Performance on Chest X-Ray}
In our study, we conducted zero-shot classification tasks on the RSNA Pneumonia, SIMM, NIH ChestXray14, and CXR-LT datasets. As depicted in Fig. \ref{fig:zeroshot_cls}a, the saliency-based methods (ReContrast and RD++) yield lower scores in comparison to the vision-language multi-modal pre-training techniques. This highlights the significance of incorporating expert-knowledge-rich text reports. MedKLIP excels in this task, likely due to its distinctive semantic simplification of text reports, facilitating effective learning of classification-relevant information. Nevertheless, our AFLoc achieves the highest AUROC scores of 0.881 (RSNA), 0.902 (SIIM), 0.737 (NIH ChestXray14), 0.735 (CXR-LT Task 2), and 0.726 (CXR-LT Task 3) on the four datasets, respectively. 
Supplementary Table\ref{extab:zeroshot_cls_cxr} provides detailed comparisons across various datasets and diseases. 
AFLoc demonstrates notable performance improvements over MedKLIP for specific diseases like atelectasis, pneumothorax, and pleural thickening. 
These results underscore the AFLoc's ability to effectively align multi-modal information, leveraging both the visual and textual features to capture complex classification patterns. 

\textbf{Diagnosis Performance on Retinal Fundus Images}
We further applied AFLoc to retinal fundus images, the results are presented in Table \ref{tab:zeroshot_cls_fundus}. The evaluation covered nine retinal diseases: Macular Degeneration (MD), Retinopathy, Myopia, Glaucoma, Congenital Optic Disc Anomalies (CODA), Retinal Arteriosclerosis (RAS), Cataract, Macular Epiretinal Membrane (MEM), and Macular Lesion (ML). While unsupervised anomaly detection methods, ReContrast and RD++, perform well on specific diseases like Myopia (0.921 AUROC) and Cataract (0.819 AUROC), their average scores are lower (0.589 and 0.520, respectively) compared to vision-language pre-training methods. Among the label free methods, AFLoc achieves the highest scores across all diseases with an average AUROC of 0.908, which is much higher than the second-best method GLoRIA (0.772). Additionally, AFLoc's performance is comparable to RETFound, which used labeled data for fine-tuning. These results highlight AFLoc's exceptional diagnostic capability and its potential for generalizing across diverse diseases.

\textbf{Diagnosis Performance on Histopathology}
The gold standard for diagnosing many diseases remains the examination of histopathology images. Supplementary Table\ref{extab:zeroshot_cls_path} and Fig. \ref{fig:zeroshot_cls}b present the comparison of balanced accuracy for various methods on the SICAPv2, WSSS4LUAD, and DHMC LUAD datasets for the zero-shot classification task. Among the compared methods, AFLoc demonstrates superior performance across all datasets, achieving the highest balanced accuracy scores of 0.512, 0.704, and 0.442 on the SICAPv2, WSSS4LUAD, and DHMC LUAD datasets, respectively. 
The consistent performance enhancement of AFLoc can be attributed to its robust multi-modal semantic alignment, which effectively captures both visual features from histopathology images and semantic context from textual descriptions.

\begin{table}[ht]
    \centering
    \caption{Comparisons of AUROC on retinal fundus datasets for the zero-shot classification task. RETFound uses human-annotated classification labels during fine-tuning.} 
    \setlength{\tabcolsep}{2pt} 
    \begin{tabular}{lccccccccccc}
        \toprule[1pt]
        \textbf{Methods} & \textbf{Label free} & \textbf{MD} & \textbf{Retinopathy} & \textbf{Myopia} & \textbf{Glaucoma} & \textbf{CODA} & \textbf{RAS} & \textbf{Cataract} & \textbf{MEM} & \textbf{ML} & \textbf{Mean} \\ \hline
        ReContrast & \ding{51} & 0.423 & 0.689 & 0.921 & 0.693 & 0.468 & 0.408 & 0.341 & 0.546 & 0.812 & 0.589 \\
        RD++ & \ding{51} & 0.569 & 0.518 & 0.291 & 0.431 & 0.634 & 0.671 & 0.819 & 0.569 & 0.176 & 0.520 \\\hline
        ConVIRT & \ding{51} & 0.421 & 0.759 & 0.870 & 0.848 & 0.854 & 0.466 & 0.424 & 0.862 & 0.698 & 0.689 \\
        GLoRIA  & \ding{51} & 0.914 & 0.592 & 0.774 & 0.926 & 0.803 & 0.636 & 0.610 & 0.834 & 0.864 & 0.772 \\ 
        AFLoc   & \ding{51} & \textbf{0.978} & \textbf{0.837} & 0.962 & \textbf{0.939} & 0.899 & 0.644 & 0.988 & \textbf{0.978} & \textbf{0.946} & \textbf{0.908} \\\hline
        RETFound  & \ding{55} & 0.935 & 0.758 & \textbf{0.971} & 0.918 & \textbf{0.900} & \textbf{0.901} & \textbf{0.995} & 0.908 & 0.883 & \textbf{0.908}\\ 
        \bottomrule[1pt]
    \end{tabular}
    \label{tab:zeroshot_cls_fundus}
        \begin{tablenotes}
        \footnotesize
        \item Bold values represent the highest performance score among the compared methods.
    \end{tablenotes}
\end{table}

\subsubsection*{Performance Enhancement via Limited Localization Annotations}

This section aims to enhance AFLoc's clinical applicability by presenting results obtained through leveraging limited labeled data for model fine-tuning on localization tasks.

As expected, AFLoc demonstrates enhanced segmentation performance with increased fine-tuning using labeled data on the SIIM dataset (Supplementary Table\ref{extab:seg_finetuning}). The Dice score is increased from 0.772 with 1\% of the data for fine-tuning to 0.809 with 100\% of the data. Furthermore, AFLoc consistently achieves higher Dice scores than all comparative methods across varying portions of labeled fine-tuning data. These results indicate that AFLoc can effectively use limited annotations to achieve enhanced segmentation performance.

The results for pathology localization on the NIH ChestXray14 dataset are depicted in Supplementary Table\ref{extab:bbox_annotation}. Consistent performance enhancement is achieved by AFLoc over the two comparative methods across different IoU thresholds. At an IoU of 0.3, AFLoc attains a mean localization accuracy of 0.62, surpassing the performance of the two comparative methods (Li et al. (0.49) and Wang et al. (0.22)). Notably, the model excels even under stricter thresholds like IoU = 0.5 and IoU = 0.7, especially for more challenging pathologies such as atelectasis and infiltration.

Extended Data Fig. \ref{exfig:seg_finetuning} presents visual examples of AFLoc’s localization performance across four datasets: JSRT, SIIM, ChestX-Det10, and Retinal Fundus. These examples cover chest X-rays with common tiny lesions and retinal hemorrhage images. The predicted regions show strong alignment with the ground-truth annotations, demonstrating AFLoc’s ability to enhance localization accuracy using only limited labeled data. 

Overall, these findings suggest that AFLoc can further benefit from fine-tuning with annotated data, achieving competitive results in both segmentation and localization tasks. This capability is crucial for broadening AFLoc's potential clinical applications.

\subsubsection*{Ablation Study of Text Granularities}

In this section, we explore the impact of incorporating text features at different levels of granularity -- word-level, sentence-level, and report-level -- on the performance of AFLoc. The results are listed in Supplementary Table\ref{extab:ablation_3level}. When using only one level of text features, the performance of AFLoc deteriorates a lot, especially for word-level and report-level features. One possible explanation could be that the word-level features lack sufficient contextual information to capture the characteristics of the pathology, whereas the report-level features may fail to capture critical diagnostic details due to their overly macroscopic perspective. On the other hand, AFLoc using sentence-level features achieves relatively higher performance, suggesting that sentence-level features can provide richer contextual information, which is important for pathology localization in medical imaging. When these different levels of text features are combined in use, clear performance enhancement is observed, and the best performance is achieved when all three levels are employed, fully supporting our hypothesis that the combination of multi-level information is crucial for improving the accuracy of pathology localization in medical images.

\subsubsection*{Ablation Study of Text Encoders}
Text encoders play a crucial role in deeply understanding medical concepts and achieving accurate semantic alignment. To investigate the impact of text encoding on pathology localization performance, we implemented the text encoder from the latest large language and vision assistant for biomedicine (LLaVA-Med \cite{li2023llavamed}). LLaVA-Med has shown improvements in open biomedical question-answering tasks. However, as demonstrated in Supplementary Table\ref{extab:text_encoder}, its performance falls short compared to BioClinicalBERT \cite{alsentzer2019publicly}. This discrepancy can be attributed to the superior performance of language models trained on domain-specific datasets (e.g., biomedical and clinical data) tailored for their respective tasks \cite{peng2019transfer}. LLaVA-Med was trained on the PMC-15M dataset, which is extracted from scientific publications in PubMed Central, whereas BioClinicalBERT was trained on electronic health records of intensive care unit patients, closely aligning with the clinical reports used in this study. This domain-specific alignment likely provides BioClinicalBERT an advantage in comprehending intricate medical terminology and improving its localization performance.


    
    
 
  

\section*{Discussion}\label{sec12}
Pathology localization and diagnosis are essential components of clinical applications in medical imaging. However, existing methods encounter various challenges, notably stemming from the scarcity of annotated data related to generalization \cite{wu2023medklip, tiu2022expert}. In this study, we introduce AFLoc, an annotation-free deep learning model designed to enhance pathology localization by using both clinical reports and imaging data in the pre-training phase.

AFLoc employs a multi-level semantic alignment strategy to enhance the medical feature representation. Given that pathologies often occupy only small regions within images, accurate localization can be challenging for global models. While some methods align global and local text-image features \cite{huang2021gloria}, they may miss critical semantic context by focusing excessively on individual words. In contrast, AFLoc adopts a more comprehensive approach by incorporating features at different levels of granularity (word-level, sentence-level, and report-level) during pre-training. This allows AFLoc to capture the necessary details for precise pathology localization.

We evaluated AFLoc’s localization and diagnostic capabilities in an annotation-free setting across three major clinical imaging modalities: chest X-rays, retinal fundus, and histopathology.
In the chest X-ray modality, AFLoc demonstrated improved localization performance compared to existing methods, such as MedKLIP and BioViL, across multiple evaluation metrics, including Intersection over Union (IoU), Dice similarity coefficient, and Contrast-to-Noise Ratio (CNR). AFLoc effectively localized a variety of thoracic pathologies, including pneumonia, atelectasis, and pleural effusion, on datasets such as RSNA Pneumonia, MS-CXR, and CheXlocalize. Notably, AFLoc showed good performance in localizing subtle and low-contrast pathologies, such as airspace opacity and atelectasis, which are often difficult to detect but crucial for early diagnosis and treatment planning. Furthermore, AFLoc surpassed the human benchmark on the CheXlocalize dataset across five key tasks, implying its potential to compete with conventional methods that depend on human-reviewed data. Additionally, AFLoc performed well on the COVID Rural dataset, which provides X-rays of patients affected with COVID-19 pneumonia, highlighting its potential for generalizing to emerging diseases. This capability is particularly important for timely and accurate disease detection, especially during global health crises like the COVID-19 pandemic.

The use of AFLoc with retinal fundus images underscores its versatility. In retinal pathology, AFLoc performed well at localizing and identifying features such as choroidal neovascularization and drusen, surpassing other methods like GLoRIA. Given the subtlety and diffuseness of retinal diseases, precise localization is particularly challenging. AFLoc's adept performance without human-annotated data highlights its potential to improve ophthalmic diagnostics, especially in settings where large-scale data annotation is not feasible.

In histopathology, AFLoc demonstrated its effectiveness in detecting cancerous tissues and abnormal structures within prostate cancer samples, achieving better diagnosis performance than specialized models like PLIP and CONCH. Histopathological analysis is essential for cancer diagnosis and prognosis, as it involves the identification of cellular-level changes indicative of cancer. AFLoc’s success in this realm indicates its proficiency in handling intricate tissue morphology and capturing fine-grained details that are crucial for cancer detection and grading. This holds particular importance in clinical contexts, where pathologists rely on image-based assessments to make critical diagnostic decisions. AFLoc’s strong performance in this area highlights its potential to support pathologists by efficiently localizing pathological regions, allowing them to focus on higher-level diagnostic tasks.

Beyond its performance across different imaging modalities, AFLoc’s ability to outperform human benchmark performance on certain tasks (e.g., detecting opacity, atelectasis, edema, consolidation, and pleural effusion) further validates its potential for clinical deployment. These results suggest that AFLoc could serve as a complementary tool to human expertise, offering clinicians an additional support system that can help mitigate diagnostic errors and improve patient care. Moreover, AFLoc’s strong localization capabilities, even in complex or subtle cases, indicate its potential to automate some aspects of the diagnostic process, potentially enhancing efficiency in clinical practice.

Furthermore, we illustrated that AFLoc’s performance could be further improved with a small amount of annotated data, demonstrating its scalability and adaptability for clinical applications. While annotation-free localization is a valuable property, fine-tuning with labeled data can elevate AFLoc's performance. This flexibility renders AFLoc fitting for diverse clinical environments, spanning well-resourced hospitals with access to large labeled datasets to smaller clinics in resource-limited settings. AFLoc's adeptness in achieving commendable results with minimal annotated data makes it a potentially practical solution for healthcare systems globally, particularly in regions where access to expert annotators is limited.

Finally, to evaluate the model’s impact on diagnostic workflow, we conducted a controlled experiment comparing diagnostic accuracy and reading time under conditions with and without AI support. The study involved two board-certified radiologists, both with more than seven years of diagnostic experience. Data were sourced from the radiologist-annotated test subset of the MIMIC-CXR dataset. A total of 100 chest X-ray images were randomly selected and evenly assigned to two groups: one interpreted by unaided radiologists (Radiologist), and the other by AI-assisted radiologists (Radiologist + AI). In the AI-assisted group, the radiologists were provided with additional information indicating the predicted disease categories and their corresponding localization. As illustrated in Fig. \ref{fig:loc_vis}, within the clinical workflow, these maps were displayed side by side with the original chest X-ray images, helping radiologists quickly focus on potential abnormalities while retaining full control over diagnostic decisions. Under both conditions, the radiologists were asked to determine the presence or absence of 13 disease categories, along with a “no findings” label, resulting in 14 binary classification tasks per image. One point was awarded for each correct classification, with a maximum score of 14 per case. As shown in Extended Data Fig. \ref{exfig:radio_ai}, the AI-assisted group achieved higher diagnostic scores, with the average score increasing from 10.80 to 11.74. Reading time was also reduced, with the average duration decreasing by 20.5\% (from 27.92 seconds to 22.20 seconds). These results indicate that the introduction of AI not only improved diagnostic accuracy but also enhanced reading efficiency, highlighting its potential as a decision-support tool in radiology.


Despite AFLoc’s encouraging performance in zero-shot generalization, several limitations remain and motivate future work. First, the model architecture can be further improved with advanced designs. For instance, a hierarchical multi-scale feature-fusion framework—one that progressively aligns extracted image features with text features at each downsampling scale—could further strengthen feature extraction. In future work, we will further explore these fusion strategies to further elevate AFLoc’s generalization capacity and diagnostic performance.

Second, AFLoc currently generates localization results conditioned on positive classification outputs, which aligns with typical clinical workflows where localization is most relevant for suspected or confirmed findings. In the future, we also must investigate the availability of localization information in cases where the classification is uncertain or incorrect. Specifically, future work will explore integrating uncertainty estimation techniques to quantify the confidence of both classification and localization outputs. This can help flag low-confidence cases for clinician review, even when a heatmap is present. In such scenarios, we plan to include visual or textual alerts indicating potential ambiguity in the prediction. Additionally, future research will consider incorporating clinical prior knowledge into the model to assist in correcting classification errors or inaccurate localizations. We also envision a reinforcement learning framework where clinician feedback on mislocalized or misclassified examples can be used to fine-tune the model, enhancing its robustness over time iteratively.

Third, while our multimodal vision–language model demonstrates promising generalizability across X-ray, retinal fundus images, and histopathology, its adaptability to additional clinical modalities remains to be thoroughly investigated. Modalities such as ultrasound and magnetic resonance imaging—as well as non-visual data types like genomics and electrocardiography—present distinct challenges due to their heterogeneous data characteristics. Incorporating modality-specific encoders may better capture cross-modal semantic alignments. Future work could explore advanced multi-way attention mechanisms or the construction of a unified multimodal representation space, thereby enabling the model to handle three or more diverse modalities, capture complex nonlinear dependencies across heterogeneous data sources.

In conclusion, AFLoc represents a promising step forward in medical image analysis by combining the strengths of multi-modal learning with annotation-free localization and diagnosis. Its strong performance across multiple modalities and its ability to assist in detecting both common and emerging diseases position it as a versatile tool with broad clinical applicability. With further refinements and integration into clinical workflows, AFLoc has the potential to improve medical diagnostics, providing healthcare professionals with AI-driven tools to enhance the accuracy and efficiency of patient care.

\section*{Methods}\label{sec13}

\subsubsection*{The Proposed AFLoc Model}\label{sec17}

\bmhead{Text Encoding}\label{sec16}
In clinical practice, medical texts may originate from various types of content, including clinical observations, diagnostic descriptions, and related information. These texts are typically composed of multiple sentences or phrases. In GLoRIA \cite{huang2021gloria}, a block tokenization technique was used, constructing complete words by aggregating multiple subwords. On the other hand, BioViL \cite{boecking2022making} employs a custom dictionary developed from multiple datasets, aiming to reduce the frequency of words being split into subwords. Treating words as independent entities may overlook the complete semantics, leading to incorrect semantic alignment between text and image. In our proposed AFLoc, text features are extracted at three levels: word-level, sentence-level, and report-level, aiming to represent report content more comprehensively and precisely through multi-granularity semantics.

Specifically, we employ BioClinicalBERT \cite{alsentzer2019publicly} as our text encoder to extract text features. When processing a medical report \( x_t \) containing \( Q \) words and \( P \) sentences, each word is tokenized into \( q_i \) tokens. The tokenizer inputs the tokenized report into the text encoder \( e_t^{I} \in \mathbb{R}^{H} \), where H is the maximum token length. The text encoder outputs \( e_t^{O} \in \mathbb{R}^{L \times H \times D} \), L is the number of layers, and \( D \) indicates the feature dimension. We take the average feature of the last 4 layers as the subword-level feature \( t_{sub} \in \mathbb{R}^{H \times D} \). Then, we aggregate features at the word, sentence, and report levels. Particularly, for a word, all its subword-level features are summarized to obtain \( Q \) word-level features \( t_w \). 
On the other hand, all corresponding subword-level features are averaged to obtain \( P \) sentence-level features \( t_s \) and one report-level feature \( t_r \).

\bmhead{Image Encoding}\label{sec18}
\deleted{We adopt a popular architecture, ResNet-50 \cite{he2016deep}, as the backbone of our image encoder \( E_v \). When processing the input image \( x_v \), we extract image features from intermediate convolutional layers of the image encoder. Specifically, shallow features are extracted from the third down-sampling stage, and deep features are from the fourth down-sampling stage. The features from the last convolutional layer of ResNet-50 are average pooled to obtain the global image features. Then, we use two \(1\times1\) convolutions and a linear layer to adjust the dimensions of the three different levels of image features to match the text feature dimensions. We obtain shallow local features \( v_s \in \mathbb{R}^{D \times M} \), deep local features \( v_d \in \mathbb{R}^{D \times \frac{M}{4}} \), and global features \( v_g \in \mathbb{R}^D \), where \( M \) denotes the number of sub-regions in shallow features, and \( D \) represents the feature dimension.}
We adopt the widely used ResNet-50 architecture \cite{he2016deep} as the backbone of our image encoder \( E_v \). Given an input image \( x_v \), we extract image features at different levels from the encoder to represent shallow local features, deep local features, and global features respectively. Specifically, shallow local features \( v_s \in \mathbb{R}^{D \times M} \) are extracted from the third down-sampling stage, while deep local features \( v_d \in \mathbb{R}^{D \times \frac{M}{4}} \) are extracted from the fourth down-sampling stage. Here, \( M \) denotes the number of sub-regions in the shallow local features, and \( D \) represents the feature dimension. The output of the final convolutional layer of the image encoder is average pooled to produce the global image features \( v_g \in \mathbb{R}^D \). To match the dimensionality of the image features with that of the text features, we applied a projection layer to all three levels of image features respectively.

\bmhead{Multi-Level Semantic Alignment}\label{sec19}
The medical report is the textual description of the corresponding medical image, containing rich information at various levels of granularity to assist in diagnosis. Ideally, the semantic information in the medical image and its corresponding report should be consistent across these different levels. This motivates us to leverage multi-level semantic information to enhance cross-modal alignment. Specifically, AFLoc learns to align the shallow local features $v_s$ and deep local features $v_d$ from the image with word-level text features $t_w$ and sentence-level text features $t_s$. In addition to local alignment, we also maintain global semantic alignment between the image's global feature $v_g$ and report-level text features $t_r$. This multi-level semantic alignment scheme is illustrated in Fig. \ref{fig:method}.

Taking the alignment between sentence feature $t_s$ and deep local feature $v_d$ as an example. The similarity matrix $s$ between $t_s$ and $v_d$ can be computed by:
\begin{equation}
s=v_d^Tt_s
\end{equation}
With the similarity $s$, the attention weighted image feature $c_i$ can be represented based on the normalized similarity for a sentence across all deep local features:
\begin{equation}
c_i= \sum_{j=1}^{\frac{M}{4}} log\frac{exp(s_{ij})/\tau _1}{ {\textstyle \sum_{k=1}^{\frac{M}{4}}} exp(s_{ik})/\tau _1}v_{dj}
\end{equation}
Then the localized feature matching function $Z\left(\cdot, \cdot\right)$ can be represented as:
\begin{equation}
Z(v,t)= log(\sum_{i=1}^{N} {exp( \Phi( c_i,t_{i}) /\tau _2}))
\end{equation}
\(\tau _1\) and \(\tau _2\) are temperature parameters, and \( N \) represents the size of the first dimension of the local features (i.e., $P$ for sentence-level features and $Q$ for word-level features). \(\Phi (c_i, t_i)\) is used to calculate the cosine similarity between the two vectors $(c_i, t_i)$. The local contrastive loss between deep local features and sentence-level features is defined as:
\begin{equation}
L_{DS} = -\frac{1}{B} \sum_{i}^{B}\left ( log\frac{exp(Z(v_d^i,t_s^i) /\tau _3)}{ {\textstyle \sum_{k=1}^{B}} exp(Z(v_d^i,t_s^k)/ \tau _3)} + log\frac{exp(Z(v_d^i,t_s^i)/\tau _3 )}{ {\textstyle \sum_{k=1}^{B}} exp(Z(v_d^k,t_s^i)/\tau _3 )} \right )
\end{equation}
where \(\tau _3\) is a temperature parameter and \(B\) is the batch size.
Similarly, we obtain the contrastive loss between shallow local features and word-level features:
\begin{equation}
L_{SW} = -\frac{1}{B} \sum_{i}^{B}\left ( log\frac{exp(Z(v_s^i,t_w^i) /\tau _3)}{ {\textstyle \sum_{k=1}^{B}} exp(Z(v_s^i,t_w^k)/ \tau _3)} + log\frac{exp(Z(v_s^i,t_w^i)/\tau _3 )}{ {\textstyle \sum_{k=1}^{B}} exp(Z(v_s^k,t_w^i)/\tau _3 )} \right )
\end{equation}
For global semantic alignment, we optimize the global features and report-level features according to the following loss function:
\begin{equation}
L_{GR} = -\frac{1}{B} \sum_{i}^{B}\left ( log\frac{exp(\Phi (v_g^i,t_r^i) /\tau _3)}{ {\textstyle \sum_{k=1}^{B}} exp(\Phi (v_g^i,t_r^k)/ \tau _3)} + log\frac{exp(\Phi (v_g^i,t_r^i)/\tau _3 )}{ {\textstyle \sum_{k=1}^{B}} exp(\Phi (v_g^k,t_r^i)/\tau _3 )} \right )
\end{equation}

AFLoc is trained to jointly optimize the local and global semantic alignment. The final loss function of AFLoc is:
\begin{equation}
L=L_{SW}+L_{DS}+L_{GR}
\end{equation}


\bmhead{Annotation-Free Pathology Localization and Diagnosis}\label{sec20}
We transform the task into a text-image matching problem using text descriptions, offering greater flexibility as the text descriptions can be easily generated to meet various clinical requirements. By default, we use the automatically generated descriptions based on the pathology’s name and a template for testing. That is, for example, `Findings suggesting \{lesion\}' for a positive description and `No findings suggesting \{lesion\}' for a negative description. We refer to such description as  `Simple description'. In addition, AFLoc also supports the `Precise description' of the pathology, such as `Large right-sided pneumothorax, air space opacity in a right infrahilar location'. This `precise description' comes from clinicians with their observation of the images. They can write corresponding descriptions to validate their hypotheses when the diagnosis is uncertain. With the text descriptions, the text features $t_r$ can be obtained with the pre-trained text encoder.

For the diagnosis task, we calculate the similarity between the global image features $v_g$ and the global text features $t_r$ for both positive and negative text descriptions. The positive similarity is then normalized to the range [0,1] as the prediction probability by softmax along the positive/negative dimension. The prediction will be considered as a positive prediction when the probability value is above a certain threshold.

For the localization task, we compute a similarity map between the deep local image features $v_d$ and the text features $t_r$ since it requires a more detailed representation. A bilinear interpolation and a normalization operation were applied to upscale the similarity map to the original image size and normalize it to the range of [-1, 1] as the generated heatmap for pathology localization. Finally, we use hard thresholding to obtain the binarized segmentation masks from the heatmaps. To ensure the reliability of the localization result, we only output the localization results with positive diagnosis predictions.

\bmhead{Experimental Details}\label{sec180}
We pre-train AFLoc on MIMIC-CXR, Retinal Fundus, and Quilt-1M dataset for chest X-ray, retinal fundus, and histopathology respectively. All images were uniformly resized to $299\times299$ pixels and we followed the same image preprocessing strategy as GLoRIA \cite{huang2021gloria}. 

A random shuffling and a sentence sampling strategy were applied to enhance the text data. Notably, the reports for the chest X-ray pre-train include multiple sections with different contents. We use the texts within the 'Findings' and 'Impressions' sections to train our model. Furthermore, we applied an additional sentence augmentation strategy to enhance the diversity of medical reports for retinal fundus data. Specifically, we generate five semantically equivalent but differently expressed sentences for each sentence by ChatGPT. These equivalent sentences will be randomly replaced during pre-training. For the histopathology dataset, we directly use the corresponding caption provided by the dataset for pre-training.

During pre-training, we used the Adam optimizer with an initial learning rate of 0.00002. A linear learning rate decay mechanism is employed by a decay factor of 0.9. The other hyperparameters were optimized with a corresponding validation dataset.

\bmhead{Evaluation Metrics}\label{sec190}
For the localization task, the Intersection over Union (IoU) and Dice coefficients were used to evaluate the thresholded results. We report the average results at multiple thresholds following the settings from BioViL. For the CheXlocalize dataset, we followed the methodology from \cite{saporta2022benchmarking} that determined an optimal threshold from the validation set and applied it to the test set. Additionally, we used the CNR \cite{boecking2022making} for measuring the difference in heatmap scores between inside and outside the bounding box area, eliminating the need for a hard threshold. This assessment of local similarity is significant, as certain clinical downstream applications might benefit more from heatmap visualizations than discrete segmentation masks. To directly assess localization performance without the influence of classification errors, we reported localization performance on the positive slices of the dataset (those with ground truth segmentations or bounding boxes annotated by radiologists). This approach ensured that the evaluation focused on the accuracy of localization, without being confounded by classification errors \cite{saporta2022benchmarking}. The averages of these metrics are reported on the test set, based on over 1,000 bootstrap repetitions, accompanied by the 95\% confidence interval (CI), which was determined using the 2.5th and 97.5th percentiles of the empirical distribution.

For the diagnosis task, we reported the area under the receiver operating characteristic curve (AUROC) and balanced accuracy. The AUROC is calculated in the binary setting to reflect the overall classification performance at different thresholds. For multi-class classification, we computed the AUROC for each disease category and then averaged them to obtain the general AUROC. Balanced accuracy is defined as the macro average of recall for each class, which accounts for the class imbalance by treating each class equally.

\subsubsection*{Dataset}\label{sec14}
\deleted{In our chest X-ray experiments, we utilized the MIMIC-CXR dataset for training. We then evaluated the model across six external datasets: RSNA Pneumonia, MS-CXR, CheXlocalize, COVID Rural, SIIM-ACR Pneumothorax (SIIM), ChestX-Det10 and CXR-LT. Additionally, to validate the model's adaptability to different modalities, we collected retinal fundus data. For the retinal fundus experiments, the model was retrained using image-report pairs from this dataset.}

We evaluated our model across three different imaging modalities: chest X-ray, retinal fundus, and histopathology. For the chest X-ray experiments, we pre-trained our model on the MIMIC-CXR dataset, then evaluated it across 8 widely used external datasets: RSNA Pneumonia (RSNA), MS-CXR, CheXlocalize, COVID Rural, SIIM-ACR Pneumothorax (SIIM), NIH ChestXray14,  ChestX-Det10, and CXR-LT. These datasets encompass 34 types of diseases, such as pneumonia, pleural effusion, and pneumothorax. For the retinal fundus experiments, we pre-trained and evaluated the model using a real-world in-house dataset. For the histopathology experiments, we pre-trained our model on the Quilt-1M dataset and then evaluated it on three external datasets: SICAPv2, WSSS4LUAD, and DHMC LUAD. The details of each dataset are provided below.

MIMIC-CXR \cite{johnson2019mimic}. The MIMIC-CXR dataset is an extensive, publicly available collection of chest X-ray images, paired with free-text radiological reports. It contains 377,110 images that are linked to 227,835 radiological studies. Each study within the database may consist of one or more images, providing a diverse range of scanning perspectives for analysis and research purposes.

RSNA Pneumonia (RSNA) \cite{shih2019augmenting}. The RSNA Pneumonia dataset presents a challenge in identifying visual signals of pneumonia from medical images. Pneumonia typically presents as one or multiple regions of increased opacity in chest X-rays \cite{franquet2018imaging}. Images in this dataset are accompanied by bounding box labels that indicate the areas affected by pneumonia. Over 1,000 cases in this dataset are positive, with each image potentially containing one or more boundary boxes highlighting the pneumonia regions.

MS-CXR \cite{boecking2022making}. MS-CXR is a curated database that features 1,153 image-sentence pairs of bounding boxes along with their corresponding phrases. This collection encompasses 8 different types of cardiopulmonary radiological findings. All the data have been annotated and validated by professional radiologists, ensuring a gold standard for phrase grounding assessment. All data in MS-CXR have been removed from the MIMIC-CXR training set to ensure it was not used in the training process.

CheXlocalize \cite{saporta2022benchmarking}. CheXlocalize is a comprehensively curated chest X-ray dataset that provides high-quality, multi-label, pixel-level segmentations. This dataset includes 234 images, encompassing 643 segmented pathologies. Moreover, CheXlocalize also provides human benchmark segmentations, drawn by a separate group of 3 board-certified radiologists, which can be used to study human benchmark localization performance.

COVID Rural \cite{desai2020chest,tang2020deep}. The COVID Rural project includes 221 chest X-ray images from 105 patients who tested positive for COVID-19. In these images, opaque areas associated with COVID-19 infection are delineated by polygons that connect specific points. The manual annotations of these images were generated by primary physicians and reviewed by experienced radiologists.

SIIM-ACR Pneumothorax (SIIM) \cite{siim}. The SIIM is a widely used pneumothorax dataset designed for the development and evaluation of detection models. In severe cases, pneumothorax can cause obstructive shock and be fatal if untreated \cite{sahn2000spontaneous}. This dataset includes more than 120,000 frontal chest X-rays, each of which has been precisely annotated. Employing the data partitioning method described by \cite{wu2023medklip}, we conducted tests on the test set.

NIH ChestXray14 \cite{wang_chestx-ray8_2017}. This dataset contains 112,120 chest X-ray images, covering 14 different potentially concurrent chest diseases. For the classification task, we use the official data split for evaluation. For the localization task, we evaluate all 8 categories with localization annotations and follow the settings of \cite{li2018thoracic} to ensure a fair comparison.

ChestX-Det10 \cite{liu2020chestxdet10}. This dataset is a subset of the NIH ChestXray14 dataset, consisting of 3,543 chest X-ray images. These images have been annotated with bounding boxes for 10 diseases/abnormalities by three board-certified radiologists. In this study, we followed the official data splits and conducted qualitative visual evaluations of our model on the test set.


CXR-LT \cite{CXR-LT,CXR-LT-MIA,PhysioNet}. This dataset is a large-scale chest X-ray dataset designed for studying multi-label and long-tailed disease classification from the CXR-LT 2024 Challenge. Compared to CXR-LT 2023 \cite{holste2023cxr}, it includes 19 additional rare disease labels. In this study, we evaluated the zero-shot classification performance of our model on the test sets, specifically for Task 2 (26 labels, including 12 newly annotated ones) and Task 3 (5 unseen labels).

Retinal Fundus. This dataset was compiled from a total of 26,028 retinal fundus images and paired reports from 11,372 individuals. All fundus images were captured using a Zeiss fundus camera and annotated by retinal experts with 10 years of experience for nine common retinal diseases. We divided the data set into training, validation, and testing sets by 8:1:1 based on the patient's unique ID, which ensures that the same patient does not appear in the same data set. Our model was trained and validated on the training and validation sets, respectively, and classification performance evaluations were done on the testing dataset. To evaluate the performance of our model in pathology localization, we used an additional dataset for validation, which was annotated at the pixel level by retinal specialists. Specifically, this dataset includes 437 images of choroidal neovascularization, 160 images of drusen, and 498 images of intraretinal hemorrhage. The study was approved by the institutional ethics committee. It was in compliance with the principles of the Declaration of Helsinki, the Good Clinical Practice (GCP), the ICH-GCP, and other relevant national laws and regulations.

Quilt-1M \cite{quilt}. This dataset contains 1 million image-text pairs for histopathology. These data were collected from diverse sources, including medical Twitter, YouTube, and other internet repositories. The dataset comprises over 437,000 unique images paired with 802,000 corresponding text annotations, encompassing a total of 1.469 million UMLS entities extracted from the text.

SICAPv2 \cite{sicap}. The SICAPv2 dataset containing prostate histology whole slide images with both annotations of global Gleason scores and patch-level Gleason grades. We use the official test set with 2,122 images from 31 whole slide images for both annotation-free localization and diagnosis tasks.

WSSS4LUAD \cite{wsss4luad,wsss4luad2}. This dataset consists of 67 hematoxylin and eosin stained whole-slide images from  Guangdong Provincial People's Hospital (GDPH) and 20 whole slide images from the cancer genome atlas (TCGA). We filtered out 4,693 image patches from the official training split following \cite{conch} for our evaluation. Each patch was labeled as tumor epithelial tissue, tumor-associated stroma tissue and normal tissue.

DHMC LUAD \cite{dhmcluad}. This dataset consists of 143 hematoxylin and eosin stained whole-slide images of lung adenocarcinoma from the Department of Pathology and Laboratory Medicine at Dartmouth-Hitchcock Medical Center (DHMC). All images were labeled into 5 categories, including lepidic, acinar, papillary, micropapillary, and solid. We use the entire dataset for annotation-free diagnosis task following \cite{conch}.

\subsubsection*{Comparative Methods}\label{sec15}
In this section, we introduced the comparative algorithms we employed, which are categorized into four main groups: vision-language pre-training methods, self-supervised pre-training methods, unsupervised anomaly detection methods, and saliency-based methods. Each category will be discussed in detail in the following subsections.

\bmhead{Medical Vision-Language Pre-Training Methods}
Ten medical vision-language pre-training methods were compared in this study, including ConVIRT \cite{convirt}, GLoRIA \cite{huang2021gloria}, BioViL \cite{boecking2022making}, MedKLIP \cite{wu2023medklip}, CLIP \cite{clip}, PLIP \cite{plip}, BiomedCLIP \cite{biomedclip}, QuiltNet \cite{quilt}, and CONCH \cite{conch}. ConVIRT improves visual representations by maximizing the agreement between true image-text pairs versus random pairs via a bidirectional contrastive objective between the image and text modalities. GLoRIA proposes an attention-based framework that learns global and local representations by comparing subregions of images with words in paired reports.
BioViL achieves better text modeling by adopting an improved lexicon and innovative language pre-training methods. MedKLIP extracts medical information from radiology reports via a triplet extraction module and uses a transformer-based fusion model to integrate this information with image features. Other models are compared as state-of-the-art algorithms for histopathology, which can be found on Hugging Face. These models represent the latest advancements in medical vision-language pre-training. 

\bmhead{Self-Supervised Pre-Training Methods}
RETFound \cite{zhou2023foundation} is a self-supervised pre-training model for retinal images that learns generalizable representations from unlabeled retinal images, providing a basis for label-efficient model adaptation across various applications. In this study, we used this model for transfer learning on the retinal fundus dataset using classification annotations and compared it with our proposed annotations-free AFLoc.

\bmhead{Unsupervised Anomaly Detection Methods}
This study compares ReContrast \cite{ReContrast} and RD++ \cite{rd++}, two state-of-the-art unsupervised anomaly detection methods. ReContrast proposes a novel contrastive learning paradigm specifically designed for domain-specific unsupervised anomaly detection. This method enhances the transferability of the pre-trained encoder by jointly optimizing all parameters in an end-to-end fashion. 
The RD++ approach improves anomaly detection by increasing feature compactness and suppressing anomalous signals through a multi-task learning design.

\bmhead{Saliency-Based Methods}
In addition to medical vision-language pre-training methods, 3 comparative saliency methods were also adopted in our study. 
Grad-CAM \cite{gradcam} uses the gradients flowing into the final convolutional layer for any target category to generate a coarse localization map, highlighting the important areas in the image used for classification. 
Grad-CAM++ \cite{gradcam++} is an extension of Grad-CAM that aims to enhance and refine the visualization of the decision-making process of convolutional neural networks. Compared to the original Grad-CAM approach, Grad-CAM++ can provide clearer and more detailed visual explanations, especially in identifying specific classes in recognition models.
Eigen-CAM \cite{eigencam} is a visualization technique to interpret the decisions of convolutional neural networks. It extends the traditional Class Activation Mapping (CAM) method, aiming to offer a more intuitive and detailed visual interpretation.

\bmhead{Disease Localization Methods}
To evaluate the potential performance improvement of our method with limited location annotations, we compared with the methods proposed by Wang et al. \cite{wang_chestx-ray8_2017} and Li et al. \cite{li2018thoracic}. Wang et al. \cite{wang_chestx-ray8_2017} proposed a weakly-supervised framework for weakly-supervised object localization by considering large image capacity, various multi-label CNN losses and different pooling strategies. Li et al. \cite{li2018thoracic} proposed a unified approach that effectively leverage class information and limited location annotations. 

\backmatter

\section*{Acknowledgments}
This research was partly supported by the National Natural Science Foundation of China (62222118 to S.S.W., U22A2040 to S.S.W., W2431057 to K.Z.); Shenzhen Medical Research Fund (B2402047 to S.S.W.); the major key project of Pengcheng Laboratory under grant PCL2024A06 (to H.Y., J.L., and W.H.); Key Laboratory for Magnetic Resonance and Multimodality Imaging of Guangdong Province (2023B1212060052 to S.S.W.); Youth Innovation Promotion Association CAS (to S.S.W.); the Macau Science and Technology Development Fund, Macao (0007/2020/AFJ to K.Z., 0070/2020/A2 to K.Z. and 0003/2021/AKP to K.Z.); Macau University of Science and Technology; State Key Laboratory for Eye Health, Eye Hospital and Institute for Advanced Study on Eye Health and Diseases, Wenzhou Medical University.

\section*{Author contributions}
S.S.W. and H.Y. conceived the idea. S.S.W., H.Y., H.-Y.Z., and K.Z. designed the experiments. H.Y., J.L., and W.H. implemented and conducted the experiments. H.Y., H.-Y.Z., J.L., W.H., and C.L. analyzed the data and experimental results. S.S.W., H.Y., H.-Y.Z., J.L., W.H., and C.L. wrote the paper. K.Z., Z.L., and Y.G. contributed the retinal fundus dataset and analysis tools. Q.L., Y.L., Q.Y., S.W., T.T., and H.Z. participated in discussions and contributed critical feedback. All authors reviewed and approved the final version of the paper.

\section*{Data Availability}
MIMIC-CXR data is available at \url{https://physionet.org/content/mimic-cxr/2.0.0} for users with credentialed access. RSNA Pneumonia data is available at \url{https://www.rsna.org/rsnai/ai-image-challenge/RSNA-Pneumonia-Detection-Challenge-2018}. MS-CXR data is available at \url{https://physionet.org/content/ms-cxr/0.1/}. CheXlocalize data is available at \url{https://stanfordaimi.azurewebsites.net/datasets/23c56a0d-15de-405b-87c8-99c30138950c}. COVID Rural data is available at \url{https://www.cancerimagingarchive.net/collection/covid-19-ar/}. SIIM-ACR Pneumothorax data is available at \url{https://www.kaggle.com/c/siim-acr-pneumothorax-segmentation}. NIH ChestXray4 data is available at \url{https://nihcc.app.box.com/v/ChestXray-NIHCC}. ChestX-Det10 data is available at \url{https://github.com/Deepwise-AILab/ChestX-Det10-Dataset}. Quilt-1M data is available at \url{https://github.com/wisdomikezogwo/quilt1m}. SICAPv2 data is available at \url{https://data.mendeley.com/datasets/9xxm58dvs3/2}. WSSS4LUAD data is available at \url{https://wsss4luad.grand-challenge.org/}. DHMC LUAD data is available at \url{https://bmirds.github.io/LungCancer/}. CXR-LT data is available at \url{https://physionet.org/content/cxr-lt-iccv-workshop-cvamd/2.0.0/}. The Retinal Fundus data were collected by the hospitals. Owing to patient-privacy considerations, they are not publicly available. Researchers may contact the corresponding authors to discuss data availability for research purposes upon reasonable request.

\section*{Code Availability}
The code is available on GitHub at \url{https://github.com/YH0517/AFLoc}.

\bibliography{sn-article}

@article{tiu2022expert,
  title={Expert-level detection of pathologies from unannotated chest X-ray images via self-supervised learning},
  author={Tiu, Ekin and Talius, Ellie and Patel, Pujan and Lanrtz, Curtis P and Ng, Andrew Y and Rajpurkar, Pranav},
  journal={Nature Biomedical Engineering},
  volume={6},
  number={12},
  pages={1399--1406},
  year={2022},
  publisher={Nature Publishing Group UK London}
}

@article{wu2023medklip,
  title={{MedKLIP}: Medical Knowledge Enhanced Language-Image Pre-Training},
  author={Wu, Chaoyi and Zhang, Xiaoman and Zhang, Ya and Wang, Yanfeng and Xie, Weidi},
  journal={Proceedings of the IEEE/CVF International Conference on Computer Vision},
  year={2023}
}

@article{zhou2023foundation,
  title={A foundation model for generalizable disease detection from retinal images},
  author={Zhou, Yukun and Chia, Mark A and Wagner, Siegfried K and Ayhan, Murat S and Williamson, Dominic J and Struyven, Robbert R and Liu, Timing and Xu, Moucheng and Lozano, Mateo G and Woodward-Court, Peter and others},
  journal={Nature},
  volume={622},
  number={7981},
  pages={156--163},
  year={2023},
  publisher={Nature Publishing Group UK London}
}

@article{
zhou2023advancing,
title={Advancing Radiograph Representation Learning with Masked Record Modeling},
author={Hong-Yu Zhou and Chenyu Lian and Liansheng Wang and Yizhou Yu},
journal={The Eleventh International Conference on Learning Representations },
year={2023}
}

@article{huang2021gloria,
  title={{GLoRIA}: A multimodal global-local representation learning framework for label-efficient medical image recognition},
  author={Huang, Shih-Cheng and Shen, Liyue and Lungren, Matthew P and Yeung, Serena},
  journal={Proceedings of the IEEE/CVF International Conference on Computer Vision},
  pages={3942--3951},
  year={2021}
}

@article{boecking2022making,
  title={Making the most of text semantics to improve biomedical vision--language processing},
  author={Boecking, Benedikt and Usuyama, Naoto and Bannur, Shruthi and Castro, Daniel C and Schwaighofer, Anton and Hyland, Stephanie and Wetscherek, Maria and Naumann, Tristan and Nori, Aditya and Alvarez-Valle, Javier and others},
  journal={European Conference on Computer Vision},
  pages={1--21},
  year={2022},
  organization={Springer}
}

@article{lee2022localization,
  title={Localization-adjusted diagnostic performance and assistance effect of a computer-aided detection system for pneumothorax and consolidation},
  author={Lee, Sun Yeop and Ha, Sangwoo and Jeon, Min Gyeong and Li, Hao and Choi, Hyunju and Kim, Hwa Pyung and Choi, Ye Ra and I, Hoseok and Jeong, Yeon Joo and Park, Yoon Ha and others},
  journal={npj Digital Medicine},
  volume={5},
  number={1},
  pages={107},
  year={2022},
  publisher={Nature Publishing Group UK London}
}

@article{cao2023large,
  title={Large-scale pancreatic cancer detection via non-contrast {CT} and deep learning},
  author={Cao, Kai and Xia, Yingda and Yao, Jiawen and Han, Xu and Lambert, Lukas and Zhang, Tingting and Tang, Wei and Jin, Gang and Jiang, Hui and Fang, Xu and others},
  journal={Nature Medicine},
  pages={1--11},
  year={2023},
  publisher={Nature Publishing Group US New York}
}

@article{song2020clinically,
  title={Clinically applicable histopathological diagnosis system for gastric cancer detection using deep learning},
  author={Song, Zhigang and Zou, Shuangmei and Zhou, Weixun and Huang, Yong and Shao, Liwei and Yuan, Jing and Gou, Xiangnan and Jin, Wei and Wang, Zhanbo and Chen, Xin and others},
  journal={Nature Communications},
  volume={11},
  number={1},
  pages={4294},
  year={2020},
  publisher={Nature Publishing Group UK London}
}

@article{leon2023hyperspectral,
  title={Hyperspectral imaging benchmark based on machine learning for intraoperative brain tumour detection},
  author={Leon, Raquel and Fabelo, Himar and Ortega, Samuel and Cruz-Guerrero, Ines A and Campos-Delgado, Daniel Ulises and Szolna, Adam and Pi{\~n}eiro, Juan F and Espino, Carlos and O’Shanahan, Aruma J and Hernandez, Maria and others},
  journal={NPJ Precision Oncology},
  volume={7},
  number={1},
  pages={119},
  year={2023},
  publisher={Nature Publishing Group UK London}
}

@article{rezatofighi2019generalized,
  title={Generalized intersection over union: A metric and a loss for bounding box regression},
  author={Rezatofighi, Hamid and Tsoi, Nathan and Gwak, JunYoung and Sadeghian, Amir and Reid, Ian and Savarese, Silvio},
  journal={Proceedings of the IEEE/CVF Conference on Computer Vision and Pattern Recognition},
  pages={658--666},
  year={2019}
}

@article{johnson2019mimic,
  title={{MIMIC-CXR}, a de-identified publicly available database of chest radiographs with free-text reports},
  author={Johnson, Alistair EW and Pollard, Tom J and Berkowitz, Seth J and Greenbaum, Nathaniel R and Lungren, Matthew P and Deng, Chih-ying and Mark, Roger G and Horng, Steven},
  journal={Scientific Data},
  volume={6},
  number={1},
  pages={317},
  year={2019},
  publisher={Nature Publishing Group UK London}
}

@article{shih2019augmenting,
  title={Augmenting the national institutes of health chest radiograph dataset with expert annotations of possible pneumonia},
  author={Shih, George and Wu, Carol C and Halabi, Safwan S and Kohli, Marc D and Prevedello, Luciano M and Cook, Tessa S and Sharma, Arjun and Amorosa, Judith K and Arteaga, Veronica and Galperin-Aizenberg, Maya and others},
  journal={Radiology: Artificial Intelligence},
  volume={1},
  number={1},
  pages={e180041},
  year={2019},
  publisher={Radiological Society of North America}
}

@article{desai2020chest,
  title={Chest imaging representing a {COVID-19} positive rural US population},
  author={Desai, Shivang and Baghal, Ahmad and Wongsurawat, Thidathip and Jenjaroenpun, Piroon and Powell, Thomas and Al-Shukri, Shaymaa and Gates, Kim and Farmer, Phillip and Rutherford, Michael and Blake, Geri and others},
  journal={Scientific Data},
  volume={7},
  number={1},
  pages={414},
  year={2020},
  publisher={Nature Publishing Group UK London}
}

@article{saporta2022benchmarking,
  title={Benchmarking saliency methods for chest {X-ray} interpretation},
  author={Saporta, Adriel and Gui, Xiaotong and Agrawal, Ashwin and Pareek, Anuj and Truong, Steven QH and Nguyen, Chanh DT and Ngo, Van-Doan and Seekins, Jayne and Blankenberg, Francis G and Ng, Andrew Y and others},
  journal={Nature Machine Intelligence},
  volume={4},
  number={10},
  pages={867--878},
  year={2022},
  publisher={Nature Publishing Group UK London}
}

@article{coudray2018classification,
  title={Classification and mutation prediction from non--small cell lung cancer histopathology images using deep learning},
  author={Coudray, Nicolas and Ocampo, Paolo Santiago and Sakellaropoulos, Theodore and Narula, Navneet and Snuderl, Matija and Feny{\"o}, David and Moreira, Andre L and Razavian, Narges and Tsirigos, Aristotelis},
  journal={Nature Medicine},
  volume={24},
  number={10},
  pages={1559--1567},
  year={2018},
  publisher={Nature Publishing Group US New York}
}

@article{wang2018development,
  title={Development and validation of a deep-learning algorithm for the detection of polyps during colonoscopy},
  author={Wang, Pu and Xiao, Xiao and Glissen Brown, Jeremy R and Berzin, Tyler M and Tu, Mengtian and Xiong, Fei and Hu, Xiao and Liu, Peixi and Song, Yan and Zhang, Di and others},
  journal={Nature Biomedical Engineering},
  volume={2},
  number={10},
  pages={741--748},
  year={2018},
  publisher={Nature Publishing Group UK London}
}

@article{campanella2019clinical,
  title={Clinical-grade computational pathology using weakly supervised deep learning on whole slide images},
  author={Campanella, Gabriele and Hanna, Matthew G and Geneslaw, Luke and Miraflor, Allen and Werneck Krauss Silva, Vitor and Busam, Klaus J and Brogi, Edi and Reuter, Victor E and Klimstra, David S and Fuchs, Thomas J},
  journal={Nature Medicine},
  volume={25},
  number={8},
  pages={1301--1309},
  year={2019},
  publisher={Nature Publishing Group US New York}
}

@article{courtiol2019deep,
  title={Deep learning-based classification of mesothelioma improves prediction of patient outcome},
  author={Courtiol, Pierre and Maussion, Charles and Moarii, Matahi and Pronier, Elodie and Pilcer, Samuel and Sefta, Meriem and Manceron, Pierre and Toldo, Sylvain and Zaslavskiy, Mikhail and Le Stang, Nolwenn and others},
  journal={Nature Medicine},
  volume={25},
  number={10},
  pages={1519--1525},
  year={2019},
  publisher={Nature Publishing Group US New York}
}

@article{gradcam,
  author={Selvaraju, Ramprasaath R. and Cogswell, Michael and Das, Abhishek and Vedantam, Ramakrishna and Parikh, Devi and Batra, Dhruv},
  journal={2017 IEEE International Conference on Computer Vision (ICCV)}, 
  title={{Grad-CAM}: Visual Explanations from Deep Networks via Gradient-Based Localization}, 
  year={2017},
  volume={},
  number={},
  pages={618-626},
  doi={10.1109/ICCV.2017.74}
}

@article{gradcam++,
  author={Chattopadhay, Aditya and Sarkar, Anirban and Howlader, Prantik and Balasubramanian, Vineeth N},
  journal={2018 IEEE Winter Conference on Applications of Computer Vision (WACV)}, 
  title={{Grad-CAM++}: Generalized Gradient-Based Visual Explanations for Deep Convolutional Networks}, 
  year={2018},
  volume={},
  number={},
  pages={839-847},
  doi={10.1109/WACV.2018.00097}}

@article{eigencam,
  title={Eigen-cam: Class activation map using principal components},
  author={Muhammad, Mohammed Bany and Yeasin, Mohammed},
  journal={2020 International Joint Conference on Neural Networks (IJCNN)},
  pages={1--7},
  year={2020},
  organization={IEEE}
}

@article{he2016deep,
  title={Deep residual learning for image recognition},
  author={He, Kaiming and Zhang, Xiangyu and Ren, Shaoqing and Sun, Jian},
  journal={Proceedings of the IEEE Conference on Computer Vision and Pattern Recognition},
  pages={770--778},
  year={2016}
}

@article{alsentzer2019publicly,
  title={Publicly Available Clinical {BERT} Embeddings},
  author={Alsentzer, Emily and Murphy, John R and Boag, Willie and Weng, Wei-Hung and Jin, Di and Naumann, Tristan and Redmond, WA and McDermott, Matthew BA},
  journal={NAACL HLT 2019},
  pages={72},
  year={2019}
}

@article{franquet2018imaging,
  title={Imaging of community-acquired pneumonia},
  author={Franquet, Tom{\'a}s},
  journal={Journal of Thoracic Imaging},
  volume={33},
  number={5},
  pages={282--294},
  year={2018},
  publisher={Wolters Kluwer}
}

@article{tang2020deep,
  title={Deep learning segmentation model for automated detection of the opacity regions in the chest {X-rays} of the {Covid-19} positive patients and the application for disease severity},
  author={Tang, Haiming and Sun, Nanfei and Li, Yi and Xia, Haoran},
  journal={medRxiv},
  pages={2020--10},
  year={2020},
  publisher={Cold Spring Harbor Laboratory Press}
}

@article{chen2020simple,
  title={A simple framework for contrastive learning of visual representations},
  author={Chen, Ting and Kornblith, Simon and Norouzi, Mohammad and Hinton, Geoffrey},
  journal={International Conference on Machine Learning},
  pages={1597--1607},
  year={2020},
  organization={PMLR}
}

@article{morens2013emerging,
  title={Emerging infectious diseases: threats to human health and global stability},
  author={Morens, David M and Fauci, Anthony S},
  journal={PLoS Pathogens},
  volume={9},
  number={7},
  pages={e1003467},
  year={2013},
  publisher={Public Library of Science San Francisco, USA}
}

@article{convirt,
  title={Contrastive learning of medical visual representations from paired images and text},
  author={Zhang, Yuhao and Jiang, Hang and Miura, Yasuhide and Manning, Christopher D and Langlotz, Curtis P},
  pages={2--25},
  year={2022},
  journal={Machine learning for healthcare conference}
}

@article{siim,
    author = {Anna, Zawacki and Carol, Wu and George, Shih and Julia, Elliott and Mikhail, Fomitchev and Mohannad, Hussain and ParasLakhani and Phil, Culliton and Shunxing Bao},
    title = {{SIIM-ACR} Pneumothorax Segmentation},
    publisher = {Kaggle},
    year = {2019},
    url = {https://kaggle.com/competitions/siim-acr-pneumothorax-segmentation}
}

@article{upd,
  author={Lagogiannis, Ioannis and Meissen, Felix and Kaissis, Georgios and Rueckert, Daniel},
  journal={IEEE Transactions on Medical Imaging}, 
  title={Unsupervised Pathology Detection: A Deep Dive Into the State of the Art}, 
  year={2024},
  volume={43},
  number={1},
  pages={241-252},
  doi={10.1109/TMI.2023.3298093}}

@article{baur2021autoencoders,
  title={Autoencoders for unsupervised anomaly segmentation in brain {MR} images: A comparative study},
  author={Baur, Christoph and Denner, Stefan and Wiestler, Benedikt and Navab, Nassir and Albarqouni, Shadi},
  journal={Medical Image Analysis},
  pages={101952},
  year={2021},
  publisher={Elsevier}
}

@article{ReContrast,
 author = {Guo, Jia and lu, shuai and Jia, Lize and Zhang, Weihang and Li, Huiqi},
 journal = {Advances in Neural Information Processing Systems},
 editor = {A. Oh and T. Neumann and A. Globerson and K. Saenko and M. Hardt and S. Levine},
 pages = {10721--10740},
 publisher = {Curran Associates, Inc.},
 title = {{ReContrast}: Domain-Specific Anomaly Detection via Contrastive Reconstruction},
 volume = {36},
 year = {2023}
}

@article{rd++,
    author    = {Tien, Tran Dinh and Nguyen, Anh Tuan and Tran, Nguyen Hoang and Huy, Ta Duc and Duong, Soan T.M. and Nguyen, Chanh D. Tr. and Truong, Steven Q. H.},
    title     = {Revisiting Reverse Distillation for Anomaly Detection},
    journal = {Proceedings of the IEEE/CVF Conference on Computer Vision and Pattern Recognition},
    month     = {June},
    year      = {2023},
    pages     = {24511-24520}
}

@article{rd,
  author={Deng, Hanqiu and Li, Xingyu},
  journal={2022 IEEE/CVF Conference on Computer Vision and Pattern Recognition}, 
  title={Anomaly Detection via Reverse Distillation from One-Class Embedding}, 
  year={2022},
  pages={9727-9736},
}

@article{liu2020chestxdet10,

  title={{ChestX-Det10}: Chest X-ray Dataset on Detection of Thoracic Abnormalities},
  author={Liu, Jingyu and Lian, Jie and Yu, Yizhou},
  journal={arXiv preprint arXiv:2006.10550},
  year={2020}
}

@article{
li2023llavamed,
  title={{LLaVA-Med}: Training a large language-and-vision assistant for biomedicine in one day},
  author={Li, Chunyuan and Wong, Cliff and Zhang, Sheng and Usuyama, Naoto and Liu, Haotian and Yang, Jianwei and Naumann, Tristan and Poon, Hoifung and Gao, Jianfeng},
  journal={Advances in Neural Information Processing Systems},
  volume={36},
  pages={28541--28564},
  year={2023}
}

@article{peng2019transfer,
  author={Yifan Peng and Shankai Yan and Zhiyong Lu},
  title={{Transfer Learning in Biomedical Natural Language Processing: An Evaluation of BERT and ELMo on Ten Benchmarking Datasets}},
  booktitle={Proceedings of the 2019 Workshop on Biomedical Natural Language Processing (BioNLP 2019)},
  year={2019},
}

@article{li2018thoracic,
  title={Thoracic disease identification and localization with limited supervision},
  author={Li, Zhe and Wang, Chong and Han, Mei and Xue, Yuan and Wei, Wei and Li, Li-Jia and Fei-Fei, Li},
  journal={Proceedings of the IEEE conference on Computer Vision and Pattern Recognition},
  pages={8290--8299},
  year={2018}
}

@article{reed2011multifocal,
  title={Multifocal Ill-Defined Opacities},
  author={Reed, James C},
  journal={Chest Radiology},
  pages={243},
  year={2011},
  publisher={Elsevier}
}

@article{woodring1996types,
  title={Types and mechanisms of pulmonary atelectasis},
  author={Woodring, John H and Reed, James C},
  journal={Journal of Thoracic Imaging},
  volume={11},
  number={2},
  pages={92--108},
  year={1996},
  publisher={LWW}
}

@article{lim2012age,
  title={Age-related macular degeneration},
  author={Lim, Laurence S and Mitchell, Paul and Seddon, Johanna M and Holz, Frank G and Wong, Tien Y},
  journal={The Lancet},
  volume={379},
  number={9827},
  pages={1728--1738},
  year={2012},
  publisher={Elsevier}
}

@article{uhler2008optic,
  title={Optic disc hemorrhages in glaucoma and ocular hypertension: implications and recommendations},
  author={Uhler, Tara A and Piltz-Seymour, Jody},
  journal={Current Opinion in Ophthalmology},
  volume={19},
  number={2},
  pages={89--94},
  year={2008},
  publisher={LWW}
}

@article{rajpurkar2022ai,
  title={{AI} in health and medicine},
  author={Rajpurkar, Pranav and Chen, Emma and Banerjee, Oishi and Topol, Eric J},
  journal={Nature Medicine},
  volume={28},
  number={1},
  pages={31--38},
  year={2022},
  publisher={Nature Publishing Group US New York}
}

@article{wang_chestx-ray8_2017,
	title = {{ChestX}-ray8: Hospital-scale Chest X-ray Database and Benchmarks on Weakly-Supervised Classification and Localization of Common Thorax Diseases},
	author = {Wang, Xiaosong and Peng, Yifan and Lu, Le and Lu, Zhiyong and Bagheri, Mohammadhadi and Summers, Ronald M.},
    journal = {2017 {IEEE} Conference on Computer Vision and Pattern Recognition ({CVPR})},
    pages = {3462--3471},
    year = {2017},
}

@article{clip,
  title={Learning transferable visual models from natural language supervision},
  author={Radford, Alec and Kim, Jong Wook and Hallacy, Chris and Ramesh, Aditya and Goh, Gabriel and Agarwal, Sandhini and Sastry, Girish and Askell, Amanda and Mishkin, Pamela and Clark, Jack and others},
  journal={International Conference on Machine Learning},
  pages={8748--8763},
  year={2021},
}

@article{plip,
    title={A visual--language foundation model for pathology image analysis using medical Twitter},
    author={Huang, Zhi and Bianchi, Federico and Yuksekgonul, Mert and Montine, Thomas J and Zou, James},
    journal={Nature Medicine},
    pages={1--10},
    year={2023},
    publisher={Nature Publishing Group US New York}
}

@article{biomedclip,
  title={A multimodal biomedical foundation model trained from fifteen million image--text pairs},
  author={Zhang, Sheng and Xu, Yanbo and Usuyama, Naoto and Xu, Hanwen and Bagga, Jaspreet and Tinn, Robert and Preston, Sam and Rao, Rajesh and Wei, Mu and Valluri, Naveen and others},
  journal={NEJM AI},
  volume={2},
  number={1},
  pages={AIoa2400640},
  year={2025},
  publisher={Massachusetts Medical Society}
}

@article{quilt,
  title={{Quilt-1M}: One million image-text pairs for histopathology},
  author={Ikezogwo, Wisdom and Seyfioglu, Saygin and Ghezloo, Fatemeh and Geva, Dylan and Sheikh Mohammed, Fatwir and Anand, Pavan Kumar and Krishna, Ranjay and Shapiro, Linda},
  journal={Advances in Neural Information Processing Systems},
  volume={36},
  year={2024}
}

@article{conch,
  title={A visual-language foundation model for computational pathology},
  author={Lu, Ming Y and Chen, Bowen and Williamson, Drew FK and Chen, Richard J and Liang, Ivy and Ding, Tong and Jaume, Guillaume and Odintsov, Igor and Le, Long Phi and Gerber, Georg and others},
  journal={Nature Medicine},
  pages={863–874},
  volume={30},
  year={2024},
  publisher={Nature Publishing Group}
}

@article{sicap,
    title = {Going deeper through the Gleason scoring scale: An automatic end-to-end system for histology prostate grading and cribriform pattern detection},
    journal = {Computer Methods and Programs in Biomedicine},
    volume = {195},
    pages = {105637},
    year = {2020},
    issn = {0169-2607},
    doi = {https://doi.org/10.1016/j.cmpb.2020.105637},
    author = {Julio Silva-Rodríguez and Adrián Colomer and María A. Sales and Rafael Molina and Valery Naranjo},
}

@article{wsss4luad,
  title={Wsss4luad: Grand challenge on weakly-supervised tissue semantic segmentation for lung adenocarcinoma},
  author={Han, Chu and Pan, Xipeng and Yan, Lixu and Lin, Huan and Li, Bingbing and Yao, Su and Lv, Shanshan and Shi, Zhenwei and Mai, Jinhai and Lin, Jiatai and others},
  journal={arXiv preprint arXiv:2204.06455},
  year={2022}
}

@article{wsss4luad2,
    title = {Multi-Layer Pseudo-Supervision for Histopathology Tissue Semantic Segmentation using Patch-level Classification Labels},
    journal = {Medical Image Analysis},
    pages = {102487},
    year = {2022},
    issn = {1361-8415},
    author = {Chu Han and Jiatai Lin and Jinhai Mai and Yi Wang and Qingling Zhang and Bingchao Zhao and Xin Chen and Xipeng Pan and Zhenwei Shi and Zeyan Xu and Su Yao and Lixu Yan and Huan Lin and Xiaomei Huang and Changhong Liang and Guoqiang Han and Zaiyi Liu}
}

@article{dhmcluad,
	title = {Pathologist-level classification of histologic patterns on resected lung adenocarcinoma slides with deep neural networks},
	volume = {9},
	rights = {2019 The Author(s)},
	issn = {2045-2322},
	url = {https://www.nature.com/articles/s41598-019-40041-7},
	doi = {10.1038/s41598-019-40041-7},
	pages = {3358},
	number = {1},
	journaltitle = {Scientific Reports},
	journal = {Scientific Reports},
	shortjournal = {Sci Rep},
	author = {Wei, Jason W. and Tafe, Laura J. and Linnik, Yevgeniy A. and Vaickus, Louis J. and Tomita, Naofumi and Hassanpour, Saeed},
	date = {2019-03-04},
    publisher = {Nature Publishing Group},
year={2019},
}

@article{sahn2000spontaneous,
  title={Spontaneous pneumothorax},
  author={Sahn, Steven A and Heffner, John E},
  journal={New England Journal of Medicine},
  volume={342},
  number={12},
  pages={868--874},
  year={2000},
  publisher={Mass Medical Soc}
}

@article{PhysioNet,
  title={{PhysioBank, PhysioToolkit, and PhysioNet}: components of a new research resource for complex physiologic signals},
  author={Goldberger, Ary L and Amaral, Luis AN and Glass, Leon and Hausdorff, Jeffrey M and Ivanov, Plamen Ch and Mark, Roger G and Mietus, Joseph E and Moody, George B and Peng, Chung-Kang and Stanley, H Eugene},
  journal={Circulation},
  volume={101},
  number={23},
  pages={e215--e220},
  year={2000},
  publisher={Lippincott Williams \& Wilkins}
}

@article{ CXR-LT-MIA,
  title={{Towards long-tailed, multi-label disease classification from chest X-ray: Overview of the CXR-LT challenge}},
  author={Holste, Gregory and Zhou, Yiliang and Wang, Song and Jaiswal, Ajay and Lin, Mingquan and Zhuge, Sherry and Yang, Yuzhe and Kim, Dongkyun and Nguyen-Mau, Trong-Hieu and Tran, Minh-Triet and others},
  journal={Medical Image Analysis},
  pages={103224},
  year={2024},
  publisher={Elsevier}
}

@article{ CXR-LT,
  title={{CXR-LT: Multi-Label Long-Tailed Classification on Chest X-Rays (version 2.0.0)}},
  author={Holste, Gregory and Wang, Song and Jaiswal, Ajay and Yang, Yuzhe and Lin, Mingquan and Peng, Yifan and Wang, Atlas},
  journal={PhysioNet},
  doi= {10.13026/ryj9-x506},
  url= {https://doi.org/10.13026/ryj9-x506},
  year={2025}
}

@article{holste2023cxr,
  title={{CXR-LT: Multi-Label Long-Tailed Classification on Chest X-Rays (version 1.1.0)}},
  author={Holste, Gregory and Wang, Song and Jaiswal, Ajay and Yang, Yuzhe and Lin, Mingquan and Peng, Yifan and Wang, Atlas},
  journal={PhysioNet},
  doi= {10.13026/c4tr-kr83},
  url= {https://doi.org/10.13026/c4tr-kr83},
  year={2023}
}

@article{wang_annotation-efficient_2021,
    title = {Annotation-efficient deep learning for automatic medical image segmentation},
    volume = {12},
    issn = {2041-1723},
    doi = {10.1038/s41467-021-26216-9},
    pages = {5915},
    number = {1},
    journal = {Nature Communications},
    author = {Wang, Shanshan and Li, Cheng and Wang, Rongpin and Liu, Zaiyi and Wang, Meiyun and Tan, Hongna and Wu, Yaping and Liu, Xinfeng and Sun, Hui and Yang, Rui and Liu, Xin and Chen, Jie and Zhou, Huihui and Ben Ayed, Ismail and Zheng, Hairong},
    date = {2021-10-08},
    langid = {english},
    year={2021}
}

@article{huang_nc_2024,
    title = {Enhancing representation in radiography-reports foundation model: a granular alignment algorithm using masked contrastive learning},
    volume = {15},
    rights = {2024 The Author(s)},
    issn = {2041-1723},
    doi = {10.1038/s41467-024-51749-0},
    shorttitle = {Enhancing representation in radiography-reports foundation model},
    pages = {7620},
    number = {1},
    journal = {Nature Communications},
    shortjournal = {Nat Commun},
    author = {Huang, Weijian and Li, Cheng and Zhou, Hong-Yu and Yang, Hao and Liu, Jiarun and Liang, Yong and Zheng, Hairong and Zhang, Shaoting and Wang, Shanshan},
    date = {2024-09-02},
    langid = {english},
    year={2024}
}

@article{huang_mia_2024,
    title = {Enhancing the vision–language foundation model with key semantic knowledge-emphasized report refinement},
    volume = {97},
    rights = {All rights reserved},
    issn = {1361-8415},
    doi = {https://doi.org/10.1016/j.media.2024.103299},
    pages = {103299},
    journal = {Medical Image Analysis},
    author = {Huang, Weijian and Li, Cheng and Yang, Hao and Liu, Jiarun and Liang, Yong and Zheng, Hairong and Wang, Shanshan},
    year = {2024},
}

@article{liu2024swin,
  title={{Swin-UMamba†: Adapting Mamba-based vision foundation models for medical image segmentation}},
  author={Liu, Jiarun and Yang, Hao and Zhou, Hong-Yu and Yu, Lequan and Liang, Yong and Yu, Yizhou and Zhang, Shaoting and Zheng, Hairong and Wang, Shanshan},
  journal = {IEEE Transactions on Medical Imaging},
  year={2024},
  publisher={IEEE}
}
\nolinenumbers
\newpage
\begin{appendices}

\captionsetup[figure]{name=Extended Data Fig.}
\section*{Appendix A}\label{secA1}



\begin{figure}[bht]
  \centering
  \centerline{\includegraphics[width=16cm]{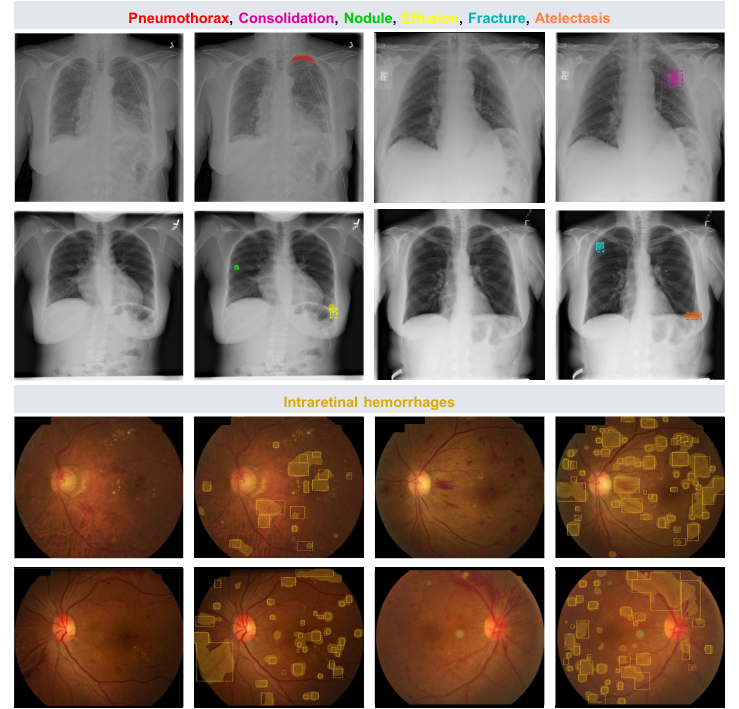}}
  \caption{Localization comparison of tiny lesions on SIIM, ChestX-Det10 and Retinal Fundus datasets. Each example shows the original image with the model's predicted lesion regions overlaid on the ground‑truth annotation. Ground‑truth boundaries are traced with dashed contours, while the model's output appears as shaded areas. Lesion types include pneumothorax, consolidation, nodule, effusion, fracture, atelectasis, and intraretinal hemorrhage.}
  \medskip
  \label{exfig:seg_finetuning}
\end{figure}

\begin{figure}[bht]
  \centering
  \centerline{\includegraphics[width=16cm]{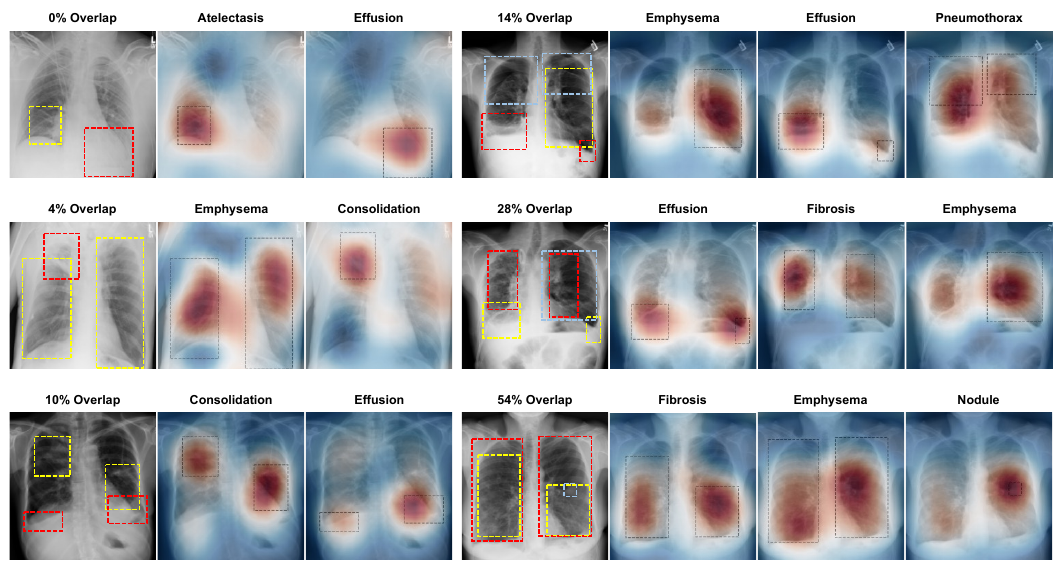}}
  \caption{Multi‑lesion localization on the ChestX‑Det10 dataset with different overlap ratios between lesions. Each example shows the original chest radiograph with annotated ground‑truth bounding boxes for co‑existing lesions (represented by dashed lines in different colors), alongside class‑specific saliency maps generated by the model based on the corresponding textual prompts. Black dotted rectangles on each saliency map highlight the ground‑truth boxes of the queried pathology. The overlap ratio is determined by calculating the overall intersection‑over‑union among all ground‑truth boxes, computed as the area of their common intersection divided by the area of their union.}
  \medskip
  \label{exfig:overlap}
\end{figure}

\begin{figure}[bht]
  \centering
  \centerline{\includegraphics[width=15cm]{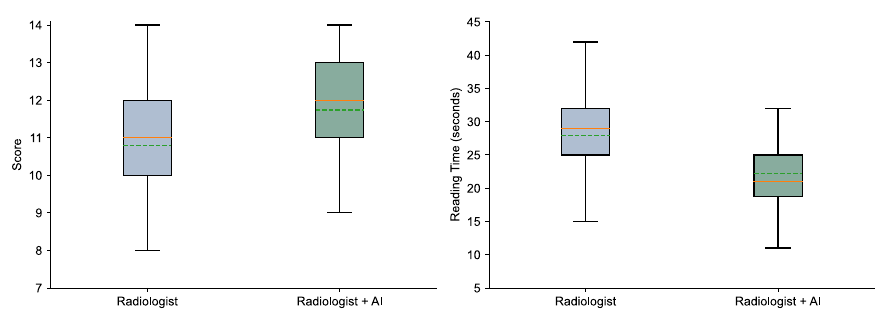}}
  \caption{Comparison of diagnostic scores and image reading times between radiologists with and without AI assistance (n = 100, with two radiologists each independently reviewing 50 images). Box plots illustrate the distribution of diagnostic scores (left) and reading times (right). In each boxplot, the solid center line represents the median, the dashed line represents the mean, the box boundaries correspond to the first and third quartiles, and the whiskers extend to the minima and maxima. Radiologists assisted by AI achieved higher diagnostic scores and reduced reading times.}
  \medskip
  \label{exfig:radio_ai}
\end{figure}
\clearpage
\section*{Supplementary Tables}
\captionsetup[table]{name=Supplementary Tables}
\begin{table}[ht]   
    \centering
    \caption{Quantitative localization results of different models on the 3 chest X-ray datasets: RSNA Pneumonia, COVID Rural, MS-CXR. Numbers within parentheses indicate 95\% CI.}
    \label{t1}
    \begin{tabular}{lccc}
        \toprule[1pt] 
        \textbf{Methods} & \textbf{RSNA Pneumonia} & \textbf{MS-CXR} & \textbf{COVID Rural} \\ \hline
        \textbf{IoU} \\ \hline
        ReContrast & {0.054 (0.050, 0.058)}  & {0.087 (0.072, 0.102)}  & {0.027 (0.020, 0.035)} \\
        RD++ & {0.050 (0.047, 0.054)}  & {0.090 (0.075, 0.105)}  & {0.058 (0.046, 0.072)} \\
        GLoRIA  & {0.278 (0.269, 0.286)}  & {0.268 (0.242, 0.295)}  & {0.138 (0.119, 0.159)} \\
        BioViL  & {0.295 (0.286, 0.304)}  & {0.228 (0.201, 0.255)}  & {0.191 (0.170, 0.215)} \\
        MedKLIP & {0.317 (0.309, 0.325)}  & {0.264 (0.240, 0.288)}  & {0.174 (0.156, 0.192)} \\
        AFLoc   & {\textbf{0.342} \textbf{(0.332, 0.350)}}  & {\textbf{0.324} \textbf{(0.298, 0.350)}}  & {\textbf{0.211} \textbf{(0.185, 0.236)}} \\ \hline
        \textbf{Dice} \\ \hline
        ReContrast & {0.095 (0.089, 0.10)}  & {0.148 (0.125, 0.171)}  & {0.049 (0.037, 0.06)} \\
        RD++ & {0.087 (0.081, 0.093)}  & {0.152 (0.130, 0.176)}  & {0.100 (0.081, 0.119)} \\
        GLoRIA  & {0.411 (0.400, 0.421)}  & {0.392 (0.357, 0.427)}  & {0.223 (0.193, 0.251)} \\
        BioViL  & {0.433 (0.423, 0.443)}  & {0.342 (0.308, 0.377)}  & {0.298 (0.268, 0.329)} \\
        MedKLIP & {0.465 (0.455, 0.474)}  & {0.394 (0.365, 0.423)}  & {0.281 (0.252, 0.308)} \\
        AFLoc   & {\textbf{0.484} \textbf{(0.474, 0.495)}}  & {\textbf{0.462} \textbf{(0.431, 0.492)}}  & {\textbf{0.323} \textbf{(0.289, 0.355)}} \\ \hline
        \textbf{CNR} \\ \hline
        ReContrast & {0.087 (0.064, 0.109)}  & {0.465 (0.398, 0.530)}  & {-0.244 (-0.325, -0.165)} \\
        RD++ & {0.225 (0.201, 0.248)}  & {0.539 (0.474, 0.602)}  & {0.289 (0.214, 0.361)} \\
        GLoRIA  & {1.223 (1.176, 1.272)}  & {1.287 (1.149, 1.421)}  & {0.663 (0.560, 0.770)}  \\
        BioViL  & {1.300 (1.262, 1.339)}  & {1.083 (0.949, 1.219)}  & {0.975 (0.880, 1.065)}  \\
        MedKLIP & {1.032 (1.014, 1.051)}  & {1.044 (0.959, 1.132)}  & {0.838 (0.772, 0.906)}  \\
        AFLoc   & {\textbf{1.538} \textbf{(1.496, 1.580)}}  & {\textbf{1.636} \textbf{(1.501, 1.772)}}  & {\textbf{1.062} \textbf{(0.929, 1.173)}}  \\
        \bottomrule[1pt]
    \end{tabular}
    \label{extab:loc_cxr3}
    \begin{tablenotes}
        \footnotesize
        \item Bold values represent the highest performance score among the compared methods.
    \end{tablenotes}
\end{table}

\begin{table}[ht]   
    \centering
    \caption{Quantitative results of localization on the retinal fundus dataset. Numbers within parentheses indicate 95\% CI.}
    \label{t1}
    \begin{tabular}{lccccc}
        \toprule[1pt] 
            \textbf{Methods}    & \makecell[c]{\textbf{Choroidal}\\\textbf{neovascularization}} & \makecell[c]{\textbf{Drusen}} & \makecell[c]{\textbf{Intraretinal}\\\textbf{hemorrhage}} & \makecell[c]{\textbf{Mean}} \\ \hline
            \textbf{IoU} \\ \hline
            ReContrast  & \makecell[c]{0.080 (0.068, 0.093)}  & \makecell[c]{0.027 (0.019, 0.035)}  & \makecell[c]{0.035 (0.031, 0.038)}  & \makecell[c]{0.047 (0.039, 0.055)}  \\
            RD++   & \makecell[c]{0.121 (0.112, 0.131)} & \makecell[c]{0.062 (0.051, 0.077)} & \makecell[c]{0.051 (0.047, 0.054)} & \makecell[c]{0.078 (0.070, 0.087)} \\
            ConVIRT  & \makecell[c]{0.131 (0.119, 0.144)}  & \makecell[c]{0.120 (0.099, 0.140)}  & \makecell[c]{0.079 (0.074, 0.085)}  & \makecell[c]{0.110 (0.097, 0.123)}  \\
            GLoRIA  & \makecell[c]{0.143 (0.130, 0.156)}  & \makecell[c]{0.107 (0.085, 0.128)}  & \makecell[c]{0.075 (0.070, 0.080)}  & \makecell[c]{0.108 (0.095, 0.121)}  \\
            AFLoc  & \makecell[c]{\textbf{0.321} \textbf{(0.303, 0.337)}}  & \makecell[c]{\textbf{0.147} \textbf{(0.125, 0.170)}}  & \makecell[c]{\textbf{0.097} \textbf{(0.091, 0.102)}}  & \makecell[c]{\textbf{0.188} \textbf{(0.173, 0.203)}} \\ \hline
            \textbf{Dice} \\ \hline
            ReContrast  & \makecell[c]{0.124 (0.107, 0.141)}  & \makecell[c]{0.046 (0.034, 0.061)}  & \makecell[c]{0.064 (0.058, 0.069)}  & \makecell[c]{0.078 (0.066, 0.090)} \\
            RD++   & \makecell[c]{0.194 (0.179, 0.209)}  & \makecell[c]{0.105 (0.087, 0.124)}  & \makecell[c]{0.092 (0.087, 0.098)}  & \makecell[c]{0.130 (0.118, 0.144)} \\
            ConVIRT  & \makecell[c]{0.205 (0.187, 0.223)}  & \makecell[c]{0.187 (0.156, 0.215)}  & \makecell[c]{0.141 (0.132, 0.149)}  & \makecell[c]{0.178 (0.158, 0.199)} \\
            GLoRIA  & \makecell[c]{0.224 (0.205, 0.242)}  & \makecell[c]{0.168 (0.137, 0.197)}  & \makecell[c]{0.134 (0.126, 0.142)}  & \makecell[c]{0.175 (0.156, 0.194)} \\
            AFLoc  & \makecell[c]{\textbf{0.451} \textbf{(0.430, 0.470)}}  & \makecell[c]{\textbf{0.225} \textbf{(0.192, 0.257)}}  & \makecell[c]{\textbf{0.170} \textbf{(0.160, 0.178)}}  & \makecell[c]{\textbf{0.282} (\textbf{0.261}, \textbf{0.302})} \\ \hline
            \textbf{CNR} \\ \hline
            ReContrast  & \makecell[c]{0.901 (0.850, 0.954)}  & \makecell[c]{0.421 (0.329, 0.510)}  & \makecell[c]{0.253 (0.231, 0.274)}  & \makecell[c]{0.525 (0.470, 0.579)}  \\
            RD++   & \makecell[c]{0.673 (0.635, 0.707)} & \makecell[c]{0.457 (0.395, 0.521)} & \makecell[c]{0.210 (0.191, 0.229)} & \makecell[c]{0.447 (0.407, 0.486)} \\
            ConVIRT  & \makecell[c]{0.359 (0.278, 0.447)}  & \makecell[c]{0.652 (0.517, 0.778)}  & \makecell[c]{0.255 (0.231, 0.278)}  & \makecell[c]{0.422 (0.342, 0.501)}  \\
            GLoRIA  & \makecell[c]{0.493 (0.427, 0.563)}  & \makecell[c]{0.655 (0.512, 0.787)}  & \makecell[c]{0.216 (0.194, 0.237)}  & \makecell[c]{0.455 (0.378, 0.529)}  \\
            AFLoc  & \makecell[c]{\textbf{1.452} \textbf{(1.368, 1.533)}}  & \makecell[c]{\textbf{0.925} \textbf{(0.783, 1.062)}}  & \makecell[c]{\textbf{0.372} \textbf{(0.349, 0.396)}}  & \makecell[c]{\textbf{0.916} \textbf{(0.833, 0.998)}}  \\
        \bottomrule[1pt]

    \end{tabular}
    \label{extab:loc_fundus}
    \begin{tablenotes}
        \footnotesize
        \item Bold values represent the highest performance score among the compared methods.
    \end{tablenotes}
\end{table}

\begin{table}[ht]
    \centering
    \caption{Quantitative localization results of different models on the SICAPv2 dataset. Numbers within parentheses indicate 95\% CI.}
    \begin{tabular}{l|ccc}
        \toprule
        \textbf{Method}     & \textbf{IoU}                  & \textbf{Dice}                 & \textbf{CNR}                   \\ \midrule
        CLIP       & 0.154 (0.141, 0.166) & 0.235 (0.218, 0.253) & 0.025 (-0.024, 0.074) \\
        BiomedCLIP & 0.188 (0.173, 0.203) & 0.279 (0.260, 0.299) & 0.078 (0.004, 0.153)  \\
        PLIP       & 0.249 (0.230, 0.269) & 0.352 (0.327, 0.376) & 0.321 (0.249, 0.398)  \\
        QuiltNet   & 0.162 (0.147, 0.177) & 0.246 (0.226, 0.265) & 0.131 (0.073, 0.188)  \\
        CONCH      & 0.198 (0.187, 0.210) & 0.300 (0.285, 0.316) & 0.378 (0.329, 0.426)  \\
        AFLoc      & \textbf{0.285 (0.267, 0.304)} & \textbf{0.399 (0.378, 0.421)} & \textbf{0.630 (0.561, 0.697)}  \\ \bottomrule
    \end{tabular}
    \label{extab:loc_path}
    \begin{tablenotes}
    \footnotesize
    \item Bold values represent the highest performance score among the compared methods.
    \end{tablenotes}
\end{table}

\begin{table}[ht]   
    \centering
    \caption{Quantitative results (IoU) of different models for different pathologies on the MS-CXR dataset. Numbers within parentheses indicate 95\% CI.}
    \setlength{\tabcolsep}{2pt} 
    \label{t1}
    \begin{tabular}{c|cccccccc}
        \toprule[1pt] 
        \textbf{Disease}           & \textbf{ReContrast}                          & \textbf{RD++}                                & \textbf{GLoRIA}                                                         & \textbf{BioViL}                              & \textbf{MedKLIP}                                                        & \textbf{AFLoc}                                                            \\ \hline
        Pneumonia        & \makecell[c]{0.090\\(0.080, 0.100)} & \makecell[c]{0.093\\(0.084, 0.103)} & \makecell[c]{0.290\\(0.265, 0.314)}                            & \makecell[c]{0.305\\(0.284, 0.328)} & \makecell[c]{0.296\\(0.277, 0.315)}                            & \makecell[c]{\textbf{0.412}\\\textbf{(0.386,} \textbf{0.435)}}   \\ 
        Atelectasis      & \makecell[c]{0.079\\(0.064, 0.095)} & \makecell[c]{0.075\\(0.063, 0.088)} & \makecell[c]{0.303\\(0.266, 0.339)}                            & \makecell[c]{0.266\\(0.222, 0.312)} & \makecell[c]{0.324\\(0.294, 0.355)}                            & \makecell[c]{\textbf{0.381}\\\textbf{(0.344,} \textbf{0.417)}}   \\ 
        Consolidation    & \makecell[c]{0.098\\(0.082, 0.115)} & \makecell[c]{0.120\\(0.101, 0.139)} & \makecell[c]{0.304\\(0.277, 0.330)}                            & \makecell[c]{0.299\\(0.273, 0.325)} & \makecell[c]{0.265\\(0.238, 0.289)}                            & \makecell[c]{\textbf{0.371}\\\textbf{(0.345,} \textbf{0.395)}}   \\ 
        Pneumothorax     & \makecell[c]{0.053\\(0.045, 0.062)} & \makecell[c]{0.093\\(0.084, 0.103)} & \makecell[c]{0.116\\(0.099, 0.133)}                            & \makecell[c]{0.109\\(0.094, 0.124)} & \makecell[c]{0.092\\(0.078, 0.106)}                            & \makecell[c]{\textbf{0.117}\\\textbf{(0.099,} \textbf{0.135)}} \\ 
        Pleural Effusion & \makecell[c]{0.061\\(0.048, 0.074)} & \makecell[c]{0.048\\(0.037, 0.058)} & \makecell[c]{0.330\\(0.304, 0.357)}                            & \makecell[c]{0.227\\(0.205, 0.250)} & \makecell[c]{0.216\\(0.195, 0.237)}                            & \makecell[c]{\textbf{0.350}\\\textbf{(0.326,} \textbf{0.375)}}   \\ 
        Lung Opacity     & \makecell[c]{0.061\\(0.046, 0.078)} & \makecell[c]{0.078\\(0.060, 0.098)} & \makecell[c]{0.196\\(0.164, 0.230)}                            & \makecell[c]{0.178\\(0.148, 0.209)} & \makecell[c]{0.197\\(0.167, 0.228)}                            & \makecell[c]{\textbf{0.327}\\\textbf{(0.290,} \textbf{0.364)}}   \\ 
        Cardiomegaly     & \makecell[c]{0.131\\(0.121, 0.140)} & \makecell[c]{0.073\\(0.066, 0.080)} & \makecell[c]{\textbf{0.408}\\\textbf{(0.398,} \textbf{0.420)}} & \makecell[c]{0.267\\(0.247, 0.286)} & \makecell[c]{0.394\\(0.384, 0.404)}                            & \makecell[c]{0.367\\(0.357, 0.377)}                              \\ 
        Edema            & \makecell[c]{0.120\\(0.092, 0.150)} & \makecell[c]{0.162\\(0.129, 0.193)} & \makecell[c]{0.201\\(0.165, 0.236)}                            & \makecell[c]{0.171\\(0.135, 0.206)} & \makecell[c]{\textbf{0.326}\\\textbf{(0.285,} \textbf{0.366)}} & \makecell[c]{0.269\\(0.237, 0.299)}                              \\ 
        \bottomrule[1pt]
    \end{tabular}
    \label{extab:loc_mscxr}
        \begin{tablenotes}
        \footnotesize
        \item Bold values represent the highest performance score among the compared methods.
    \end{tablenotes}
\end{table}

\begin{table}[]
    \centering
    \caption{Detailed comparisons of AUROC on RSNA Pneumonia, SIIM, NIH ChestXray14, and CXR-LT datasets for the zero-shot classification task. Numbers within parentheses indicate 95\% CI.}
    \begin{tabular}{c|l|ccccccccc}
        \toprule[1pt]
        \textbf{Dataset}   & \textbf{Disease}   & \rotatebox{270}{\textbf{ReContrast}} & \rotatebox{270}{\textbf{RD++}} & \rotatebox{270}{\textbf{ConVIRT}} & \rotatebox{270}{\textbf{GLoRIA}} & \rotatebox{270}{\textbf{BioViL}} & \rotatebox{270}{\textbf{CheXzero}} & \rotatebox{270}{\textbf{MedKLIP}} & \rotatebox{270}{\textbf{AFLoc}} \\ \midrule
        RSNA Pneumonia & Pneumonia          & 0.632               & 0.566         & 0.804            & 0.715           & 0.828           & 0.858             & 0.869            & \textbf{0.881 (0.868, 0.893)} \\ \midrule
        SIIM  & Pneumothorax       & 0.566               & 0.492         & 0.643            & 0.534           & 0.708           & 0.688             & 0.892            & \textbf{0.902 (0.890, 0.915)}\\ \midrule
        \multirow{15}{*}{\rotatebox{270}{NIH ChestXray14}} & Mean               & 0.511               & 0.486         & 0.560            & 0.610           & 0.662           & 0.679             & 0.726            & \textbf{0.737 (0.732, 0.741)} \\ \cmidrule(l){2-10}
              & Atelectasis        & 0.510               & 0.521         & 0.459            & 0.653           & 0.517           & 0.624             & 0.671            & \textbf{0.720 (0.710, 0.728)} \\
              & Cardiomegaly       & 0.514               & 0.485         & 0.433            & 0.704           & 0.688           & 0.837             & 0.842            & \textbf{0.856 (0.845, 0.866)} \\
              & Effuion           & 0.530               & 0.510         & 0.646            & 0.762           & 0.743           & 0.787             & 0.813   & 0.798 (0.791, 0.805)         \\
              & Infiltration       & 0.553               & 0.502         & 0.654            & 0.660           & 0.601           & 0.651             & 0.706   & 0.684 (0.676, 0.691)         \\
              & Mass               & 0.453               & 0.481         & 0.601            & 0.613           & 0.663           & 0.686             & 0.742   & 0.733 (0.720, 0.744)         \\
              & Nodule             & 0.460               & 0.482         & 0.580            & 0.508           & 0.639  & 0.580             & 0.621            & 0.617 (0.602, 0.632)         \\
              & Pneumonia          & 0.522               & 0.480         & 0.640            & 0.587           & 0.669           & 0.691             & 0.698            & \textbf{0.701 (0.681, 0.722)} \\
              & Pneumothorax       & 0.535               & 0.445         & 0.533            & 0.572           & 0.683           & 0.793             & 0.821            & \textbf{0.843 (0.835, 0.851)} \\
              & Consolidation      & 0.556               & 0.508         & 0.646            & 0.697           & 0.650           & 0.691             & 0.719            & \textbf{0.720 (0.708, 0.730)} \\
              & Edema              & 0.541               & 0.542         & 0.692            & 0.762           & 0.795           & 0.804             & 0.803            & \textbf{0.810 (0.798, 0.822)} \\
              & Emphysema          & 0.534               & 0.496         & 0.431            & 0.499           & 0.656           & 0.431             & 0.783   & 0.640 (0.627, 0.655)         \\
              & Fibrosis           & 0.466               & 0.441         & 0.482            & 0.459           & 0.632  & 0.617             & 0.604            & 0.625 (0.598, 0.655)         \\
              & Pleural thickening & 0.491               & 0.499         & 0.545            & 0.613           & 0.637           & 0.524             & 0.499            & \textbf{0.718 (0.704, 0.733)} \\
              & Hernia             & 0.485               & 0.417         & 0.494            & 0.450           & 0.698           & 0.783             & 0.841            & \textbf{0.847 (0.798, 0.884)} \\  \midrule
        \multirow{2}{*}{CXR-LT} & Task 2               & 0.526               & 0.500         & 0.643            & 0.650           & 0.648           & 0.669             & 0.677            & \textbf{0.735 (0.715, 0.756)} \\ 
        & Task 3               & 0.560               & 0.504         & 0.650            & 0.611           & 0.585           & 0.638             & 0.588            & \textbf{0.726 (0.716, 0.736)} \\ \bottomrule[1pt]
    \end{tabular}
    \label{extab:zeroshot_cls_cxr}
        \begin{tablenotes}
        \footnotesize
        \item Bold values represent the highest performance score among the compared methods.
    \end{tablenotes}
\end{table}

\begin{table}[ht]
    \centering
    \caption{Comparisons of balanced accuracy on histopathology datasets for the zero-shot classification task. Numbers within parentheses indicate 95\% CI. *: Result obtained from \cite{conch}.}
    \begin{tabular}{l|ccc}
        \toprule
        \textbf{Method}     & \textbf{SICAPv2}              & \textbf{WSSS4LUAD}            & \textbf{DHMC LUAD}            \\ \midrule
        CLIP*       & 0.250 (0.250, 0.256) & 0.333 (0.328, 0.333) & 0.194 (0.188, 0.205) \\
        BiomedCLIP* & 0.375 (0.351, 0.389) & 0.512 (0.474, 0.550) & 0.253 (0.240, 0.273) \\
        PLIP*       & 0.287 (0.268, 0.324) & 0.462 (0.439, 0.492) & 0.231 (0.222, 0.246) \\
        QuiltNet   & 0.249 (0.226, 0.327) & 0.445 (0.429, 0.461) & 0.208 (0.089, 0.347) \\
        CONCH*      & 0.339 (0.325, 0.396) & 0.598 (0.560, 0.650) & 0.314 (0.291, 0.346) \\
        AFLoc      & \textbf{0.512 (0.490, 0.533)} & \textbf{0.704 (0.692, 0.717)} & \textbf{0.442 (0.325, 0.521)}  \\ \bottomrule
    \end{tabular}
    \label{extab:zeroshot_cls_path}
        \begin{tablenotes}
        \footnotesize
        \item Bold values represent the highest performance score among the compared methods.
    \end{tablenotes}
\end{table}

\begin{table}[]
\centering
\caption{Comparisons of Dice scores with state-of-the-art methods on the fine-tuning segmentation task. Performance variations are reported for different data amounts using 1\%, 10\%, and 100\% of the data.}
\begin{tabular}{c|ccccc}
    \toprule
    \textbf{Data Portion}    & \textbf{ConVIRT}       & \textbf{GLoRIA}        & \textbf{BioViL}    & \textbf{MedKLIP}   & \textbf{AFLoc} \\\midrule
    1\%              & 0.541         & 0.567         & 0.627     & 0.666    & \textbf{0.772} \\ 
    10\%             & 0.612         & 0.578         & 0.700     & 0.721    & \textbf{0.781} \\ 
    100\%            & 0.735         & 0.769         & 0.785     & 0.794    & \textbf{0.809} \\ \bottomrule

    \end{tabular}
    \label{extab:seg_finetuning}
        \begin{tablenotes}
        \footnotesize
        \item Bold values represent the highest performance score among the compared methods.
    \end{tablenotes}
\end{table}

\begin{table}[ht]
    \centering
    \caption{Quantitative results of disease localization accuracy at various T(IoU) thresholds for eight pathologies in the NIH ChestXray14 dataset. Compared results were obtained from \cite{li2018thoracic}.}
    \begin{tabular}{c|l|cccccccc|c}
        \toprule[1pt]
        \textbf{T(IoU)}               & \textbf{Method}                                   & \rotatebox{270}{\textbf{Atelectasis}} & \rotatebox{270}{\textbf{Cardiomegaly}} & \rotatebox{270}{\textbf{Effusion}} & \rotatebox{270}{\textbf{Infiltration}} & \rotatebox{270}{\textbf{Mass}} & \rotatebox{270}{\textbf{Nodule}} & \rotatebox{270}{\textbf{Pneumonia}} & \rotatebox{270}{\textbf{Pneumothorax}} & \rotatebox{0}{\textbf{Mean}} \\ \midrule
        \multirow{3}{*}{0.1} & Wang et al. \cite{wang_chestx-ray8_2017} & 0.69                        & 0.94                         & 0.66                     & 0.71                         & 0.40                 & 0.14                   & 0.63                      & 0.38                         & 0.57                 \\
                             & Li et al. \cite{li2018thoracic}          & 0.71                        & 0.98                         & 0.87                     & 0.92                         & 0.71                 & 0.40                   & 0.60                      & 0.63                         & 0.73                 \\
                             & AFLoc                                    & 0.72                        & 1.00                         & 0.86                     & 0.91                         & 0.78                 & 0.68                   & 0.90                      & 0.82                         & \textbf{0.83}                 \\ \midrule
        \multirow{3}{*}{0.2} & Wang et al. \cite{wang_chestx-ray8_2017} & 0.47                        & 0.68                         & 0.45                     & 0.48                         & 0.26                 & 0.05                   & 0.35                      & 0.23                         & 0.37                 \\
                             & Li et al. \cite{li2018thoracic}          & 0.53                        & 0.97                         & 0.76                     & 0.83                         & 0.59                 & 0.29                   & 0.50                      & 0.51                         & 0.62                 \\
                             & AFLoc                                    & 0.59                        & 1.00                         & 0.69                     & 0.86                         & 0.66                 & 0.57                   & 0.78                      & 0.65                         & \textbf{0.73}                 \\ \midrule
        \multirow{3}{*}{0.3} & Wang et al. \cite{wang_chestx-ray8_2017} & 0.24                        & 0.46                         & 0.30                     & 0.28                         & 0.15                 & 0.04                   & 0.17                      & 0.13                         & 0.22                 \\
                             & Li et al. \cite{li2018thoracic}          & 0.36                        & 0.94                         & 0.56                     & 0.66                         & 0.45                 & 0.17                   & 0.39                      & 0.44                         & 0.49                 \\
                             & AFLoc                                    & 0.43                        & 1.00                         & 0.49                     & 0.72                         & 0.57                 & 0.52                   & 0.73                      & 0.45                         & \textbf{0.62}                 \\ \midrule
        \multirow{3}{*}{0.4} & Wang et al. \cite{wang_chestx-ray8_2017} & 0.09                        & 0.28                         & 0.20                     & 0.12                         & 0.07                 & 0.01                   & 0.08                      & 0.07                         & 0.12                 \\
                             & Li et al. \cite{li2018thoracic}          & 0.25                        & 0.88                         & 0.37                     & 0.50                         & 0.33                 & 0.11                   & 0.26                      & 0.29                         & 0.42                 \\
                             & AFLoc                                    & 0.30                        & 1.00                         & 0.26                     & 0.52                         & 0.46                 & 0.34                   & 0.57                      & 0.31                         & \textbf{0.47}                 \\ \midrule
        \multirow{3}{*}{0.5} & Wang et al. \cite{wang_chestx-ray8_2017} & 0.05                        & 0.18                         & 0.11                     & 0.07                         & 0.01                 & 0.01                   & 0.03                      & 0.03                         & 0.06                 \\
                             & Li et al. \cite{li2018thoracic}          & 0.14                        & 0.84                         & 0.22                     & 0.30                         & 0.22                 & 0.07                   & 0.17                      & 0.19                         & 0.27                 \\
                             & AFLoc                                    & 0.19                        & 0.99                         & 0.12                     & 0.32                         & 0.33                 & 0.25                   & 0.38                      & 0.21                         & \textbf{0.35}                 \\ \midrule
        \multirow{3}{*}{0.6} & Wang et al. \cite{wang_chestx-ray8_2017} & 0.02                        & 0.08                         & 0.05                     & 0.02                         & 0.00                 & 0.01                   & 0.02                      & 0.03                         & 0.03                 \\
                             & Li et al. \cite{li2018thoracic}          & 0.07                        & 0.73                         & 0.15                     & 0.18                         & 0.16                 & 0.03                   & 0.10                      & 0.12                         & 0.19                 \\
                             & AFLoc                                    & 0.09                        & 0.94                         & 0.05                     & 0.21                         & 0.22                 & 0.19                   & 0.20                      & 0.11                         & \textbf{0.25}                 \\ \midrule
        \multirow{3}{*}{0.7} & Wang et al. \cite{wang_chestx-ray8_2017} & 0.01                        & 0.03                         & 0.02                     & 0.00                         & 0.00                 & 0.00                   & 0.01                      & 0.02                         & 0.01                 \\
                             & Li et al. \cite{li2018thoracic}          & 0.04                        & 0.52                         & 0.07                     & 0.09                         & 0.11                 & 0.01                   & 0.05                      & 0.05                         & 0.12                 \\
                             & AFLoc                                    & 0.03                        & 0.78                         & 0.02                     & 0.07                         & 0.14                 & 0.08                   & 0.07                      & 0.02                         & \textbf{0.15}                 \\ \bottomrule[1pt]
    \end{tabular}
    \label{extab:bbox_annotation}
        \begin{tablenotes}
        \footnotesize
        \item Bold values represent the highest performance score among the compared methods.
    \end{tablenotes}
\end{table}

\begin{table}[ht]
    \centering
    \caption{Quantitative results (IoU) from the ablation study to investigate the importance of aligning image features with text features at different levels.}
    \setlength{\tabcolsep}{4pt} 
    \begin{tabular}{cccccccc}
        \toprule[1pt]
        \multicolumn{3}{c}{\textbf{Different Levels of Features}} & \multirow{2}{*}{\makecell[c]{\textbf{RSNA} \\\textbf{Pneumonia}}} & \multirow{2}{*}{\makecell[c]{\textbf{COVID} \\\textbf{Rural}}} & \multirow{2}{*}{\textbf{MS-CXR}} & \multirow{2}{*}{\textbf{CheXlocalize}} & \multirow{2}{*}{\textbf{Mean}} \\
        \cmidrule(r){1-3}
        \makecell[c]{\textbf{Word}--\\\textbf{Local (Shallow)}} & \makecell[c]{\textbf{Sentence}--\\\textbf{Local (Deep)}} & \makecell[c]{\textbf{Report}--\\\textbf{Global}} \\ \hline
        \ding{51} & ~ & ~ & 0.239  & 0.147  & 0.180  & 0.133  & 0.138   \\ 
        ~ & \ding{51} & ~ & 0.281  & 0.169  & 0.318  & 0.301  & 0.272   \\ 
        ~ & ~ & \ding{51} & 0.191  & 0.056  & 0.105  & 0.186  & 0.145   \\ 
        \ding{51} & \ding{51} & ~ & 0.329  & 0.177  & 0.264  & 0.285  & 0.271   \\ 
        \ding{51} & ~ & \ding{51} & 0.254  & 0.159  & 0.175  & 0.200  & 0.179   \\ 
        ~ & \ding{51} & \ding{51} & 0.285  & 0.211  & \textbf{0.325}  & 0.305  & 0.281   \\ 
        \ding{51} & \ding{51} & \ding{51} & \textbf{0.342}  & \textbf{0.211}  & 0.324  & \textbf{0.318}  & \textbf{0.299}   \\ 
        
        \bottomrule[1pt]
    \end{tabular}
    \label{extab:ablation_3level}
        \begin{tablenotes}
        \footnotesize
        \item Bold values represent the highest performance score among the compared methods.
    \end{tablenotes}
\end{table}

\begin{table}[ht]   
    \centering
    \caption{Quantitative results from replacing different text encoders in the MS-CXR dataset to investigate their impact on localization outcomes.}
    \label{t1}
    \begin{tabular}{l|ccc}
        \toprule[1pt] 
        \textbf{Text encoder} & \textbf{IoU} & \textbf{Dice} & \textbf{CNR} \\ \hline
        LLaVA-Med-v1.5-Mistral-7B & 0.307 & 0.441 & 1.483 \\
        BioClinicalBERT & \textbf{0.324} & \textbf{0.462} & \textbf{1.636} \\
        \bottomrule[1pt]
    \end{tabular}
    \label{extab:text_encoder}
        \begin{tablenotes}
        \footnotesize
        \item Bold values represent the highest performance score among the compared methods.
    \end{tablenotes}
\end{table}

\end{appendices}


\end{document}